\documentclass[acmsmall]{acmart}

\usepackage[linesnumbered,ruled,vlined]{algorithm2e}
\usepackage{float} 
\usepackage{amsmath}
\usepackage[english]{babel}
\usepackage{fancyhdr}
\usepackage{graphicx}
\usepackage[export]{adjustbox}
\usepackage{subcaption}
\usepackage{multicol}

\def\BibTeX{{\rm B\kern-.05em{\sc i\kern-.025em b}\kern-.08emT\kern-.1667em\lower.7ex\hbox{E}\kern-.125emX}}
    
\begin{document}

%
\title{Socially-Aware Navigation: A Non-linear Multi-Objective Optimization Approach}

%
\author{Santosh Balajee Banisetty}
\email{santoshbanisetty@nevada.unr.edu}
\affiliation{%
  \institution{Dept.
of Computer Science and Engineering, University of Nevada Reno}
  \streetaddress{1664 N. Virginia Street}
  \city{Reno}
  \state{Nevada}
  \postcode{89557}
}
\author{Scott Forer}
\email{sforer@nevada.unr.edu}
\affiliation{%
  \institution{Dept.
of Mechanical Engineering, University of Nevada Reno}
  \streetaddress{1664 N. Virginia Street}
  \city{Reno}
  \state{Nevada}
  \postcode{89557}
}
\author{Logan Yliniemi}
\authornote{L. Yliniemi is currently with Amazon Robotics. This work was completed prior to joining Amazon.}
\email{logan@unr.edu}
\affiliation{%
  \institution{Dept.
of Mechanical Engineering, University of Nevada Reno}
  \streetaddress{1664 N. Virginia Street}
  \city{Reno}
  \state{Nevada}
  \postcode{89557}
}
\author{Monica Nicolescu}
\email{monica@unr.edu}
\affiliation{%
  \institution{Dept.
of Computer Science and Engineering, University of Nevada Reno}
  \streetaddress{1664 N. Virginia Street}
  \city{Reno}
  \state{Nevada}
  \postcode{89557}
}
\author{David~Feil-Seifer}
\email{dave@cse.unr.edu}
\affiliation{%
  \institution{Dept.
of Computer Science and Engineering, University of Nevada Reno}
  \streetaddress{1664 N. Virginia Street}
  \city{Reno}
  \state{Nevada}
  \postcode{89557}
}

%
\renewcommand{\shortauthors}{Banisetty, et al.}

%
\begin{abstract}
Mobile robots are increasingly populating homes, hospitals, shopping malls, factory floors, and other human environments. Human society has social norms that people mutually accept; obeying these norms is an essential signal that someone is participating socially with respect to the rest of the population. For robots to be socially compatible with humans, it is crucial for robots to obey these social norms. In prior work, we demonstrated a Socially-Aware Navigation (SAN) planner, based on Pareto Concavity Elimination Transformation (PaCcET), in a hallway scenario, optimizing two objectives so that the robot does not invade the personal space of people. This paper extends our PaCcET based SAN planner to multiple scenarios with more than two objectives. We modified the Robot Operating System's (ROS) navigation stack to include PaCcET in the local planning task. We show that our approach can accommodate multiple Human-Robot Interaction (HRI) scenarios. Using the proposed approach, we achieved successful HRI in multiple scenarios like hallway interactions, an art gallery, waiting in a queue, and interacting with a group. We implemented our method on a simulated PR2 robot in a 2D simulator (Stage) and a pioneer-3DX mobile robot in the real-world to validate all the scenarios. A comprehensive set of experiments shows that our approach can handle multiple interaction scenarios on both holonomic and non-holonomic robots; hence, it can be a viable option for a Unified Socially-Aware Navigation (USAN).
\end{abstract}

%
%


%

%

%
\maketitle

\section{Introduction}
Social norms such as driving on the right or left side of the road (depending on the country one lives in), turn-taking rules at four-way stops and roundabouts, holding doors for people behind us, and maintaining an appropriate distance when interacting with another person (actual distance depending on the type of interaction) are crucial in our day-to-day interactions. People use these actions as signals that they are participants in the social order. Violating these principles is jarring at best (i.e., a person becoming confused at another person's behavior) and can provoke hostility at worst (i.e., getting upset at someone for cutting line). 

As socially assistive robots (SAR) \cite{feil2005defining} are expected to play an essential role in a human-robot collaborative environment, these robots taking roles while remaining unconstrained to such social norms is a growing concern in the robotics community. In the last decade, social and personal robots have attained immense interest among researchers and entrepreneurs alike; the result is an increase in the efforts both in industry and academia to develop applications and businesses in the personal robotics space. Smart Luggage \cite{Smartlauggage}, showcased at 2018's consumer electronics show (CES), a robotic suitcase that will follow the owner during travel, demonstrates the commercial viability of interpersonal navigation. Some companies and start-ups develop robotic assistants for airports and shopping malls to assist people with directions and shopping experiences. Robot domains, especially SAR, benefit from navigation as such movement extends the robot's reachable service area. However, navigation, if not appropriately performed, can cause an adverse social reaction~\cite{mutlu2008robots}. In an ethnographic study where nurses in hospital settings interacted with an autonomous service robot, one of the participants quoted the following statement: \hfill \break
\newline``\textit{Well; it almost ran me over... I wasn't scared... I was just mad... I've already been clipped by it. It does hurt.}''\newline

Robots currently deployed in human environments have prompted adverse reactions from people encountering them~\cite{Carlson2019}. Some people are unwilling to interact with these robots, and some even kicked them, demonstrating a very hostile attitude towards service robots. Incidents such as these pose both challenges and opportunities for the human-robot interaction (HRI) researchers; as a result, recent years have seen a tremendous growth of publication in areas related to socially-aware navigation such as human detection and tracking, social planner incorporating social norms, human-robot interaction studies to understand the problem from a human's perspective.

\begin{figure}[t]
\centering
\includegraphics[height = 4.5cm]{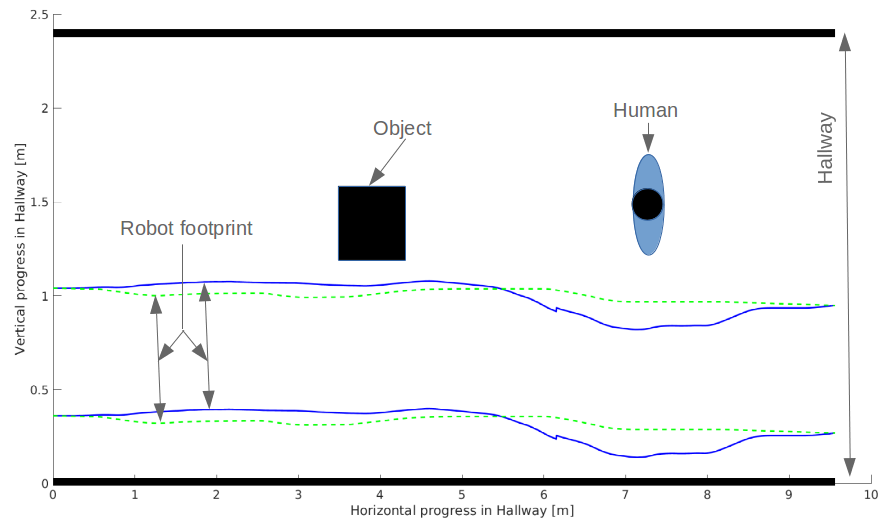}
\caption{The PaCcET local planner (blue, solid) compared with the traditional ROS local planner (green, dotted), which does not account for social norms. The traditional planner generated a trajectory that is close to both human and the object (black box), treating them alike. Our approach, the PaCcET-based SAN planner, generated a trajectory that diverged around the human, thereby respecting the personal space of the human.}
\label{fig:results}
\end{figure}

As seen from the examples so far \cite{mutlu2008robots, businessinsider}, SAR systems without social-awareness can cause problems in human environments. Following human societal social norms may help robots be more accepted in homes, hospitals, and workplaces. Robots should not treat humans as dynamic objects through their navigation behavior, as shown in our results in Figure \ref{fig:results}. The figure illustrates a comparison between a traditional planner (green trajectory) that optimizes performance (time, distance, etc.) and a socially-aware planner (blue trajectory), which optimizes for social norms along with performance metrics. The traditional planner treats both humans and objects the same way (not maintaining enough distance around humans and objects); however, that is not acceptable. It is acceptable to get close to an object, but a similar behavior around a person is not acceptable. 

Socially-Aware Navigation planners, including our method, considers the theory of proxemics and other social norms such that the robot does not invade the space around the human. Proxemics \cite{hall1966hidden} codifies this notion of personal space; researchers interested in socially-aware navigation (SAN) are investigating methods to integrate the rules of proxemics into robot navigation behavior.

Our approach to SAN using PaCcET addresses the following limitations of the existing approaches:
\begin{enumerate}
  \item Many current approaches depend on exocentric sensing, limiting the robot's services to a particular environment~\cite{feil2012distance}. 
  \item Approaches may require a large amount of training data~\cite{okal2016learning, kim2016socially, kretzschmar2016socially, alahi2016social, hamandi2018deepmotion}. 
  \item The environment/scenario is a singleton, i.e., Only a hallway, a room, etc., is considered or considers only an approach behavior or a passing behavior~\cite{feil2012distance}.
  \item Planners are optimized for single or few objectives with linear combination or weighted sum~\cite{ferrer2013robot, ferrer2017robot}. 
\end{enumerate}


This paper builds on our previous work from~\cite{forer2018socially} and provides a comprehensive evaluation of our SAN planner in multiple scenarios in both simulations and the real-world. We extend our prior work by implementing our proposed method on a real robot with objectives like inter-personal distance, adherence to a social goal, activity space, and group proxemics and providing a more diverse set of simulated environments real-world scenarios for evaluation. The remainder of this paper is structured as follows. We implemented our proposed SAN method on both holonomic and non-holonomic platforms. In the next section, we review related works. In Section \ref{sec:method}, we present our approach to SAN using PaCcET. In Section \ref{sec:results}, we apply our method to various scenarios in simulation and on a pioneer-3DX platform to validate the proposed approach. Finally, in Section \ref{sec:discussion}, we discuss our present and on-going work.    

\section{Related Work}
\label{sec:related_works}

\subsection{Socially-Aware Navigation}
When localization and navigation were relatively new in the field of robotics, tour guide robots Rhino and Minerva \cite{burgard1998interactive, burgard1999experiences, thrun1999minerva} successfully navigated and gave museum tours for visitors. These robots were some of the pioneering works in the field of robot navigation, which performed various navigation functions like mapping, localization, collision avoidance, and path planning. These robots also exhibited primitive navigation behaviors around people in a dynamic human environment. Robust long-term navigation in the indoor environment was demonstrated by Marder \textit{et al.} \cite{marder2010office} when a PR2, a mobile manipulation robot, completed a 26.2 mile run in an office environment. This state-of-the-art navigation technique (traditional planner) can generate a collision-free path and maneuver a robot on that path to get to a goal. However, these algorithms are not sophisticated enough to deal with social interactions while navigating in highly dynamic human environments.  

There is a rapidly growing HRI community that is addressing social navigation planners~\cite{truong2016dynamic, gomez2014fast, rios2011understanding, sebastian2017socially-aware, feil2012distance, kruse2014evaluating}. Most of the SAN research can be broadly classified (based on areas concerning social navigation) into Planning, Perception, and Behavior Selection. Most of the work is concentrated in the planning area (in turn classified into local planning and global planning). The solutions to SAN-associated problems range from simple cost functions to more advanced deep neural networks; we present some of them here. 

Social Force Model (SFM)~\cite{helbing1995social} is one of the popular approaches to SAN, which mimics human navigation behavior. Building upon prior work, Ferrer \textit{et al.} programmed a robot to obey the social forces during navigation activities~\cite{ferrer2013robot, ferrer2017robot}. The method also extends SFM to allow a robot to accompany a human while providing a method for learning the model parameters. Kivrak \textit{et al.} \cite{kivrak2018social} also adopted SFM to be used as a local planner to generate a socially aware trajectory in a hallway scenario. Silva \textit{et al.} \cite{silva2017human} took a shared effort approach to solve the human-robot collision avoidance problem using Reinforcement Learning. Simulated results show that the approach enabled the agents (human and robot) to avoid collisions mutually. Our proposed approach is validated not in a single context but multiple contexts like hallway interactions, joining a group of people, waiting in a line, and an art gallery interaction. 

Dondrup \textit{et al.} \cite{dondrup2016qualitative} proposed a combination of well-known sample-based planning and velocity costmaps to achieve socially-aware navigation. The authors used a Bayesian temporal model to represent the navigation intent of robot and human based on Qualitative Trajectory Calculus and used these descriptors as constraints for trajectory generation. Alonso \textit{et al.} \cite{alonso2018cooperative} presented a $\epsilon$-cooperative collision avoidance method in dynamic environments among interactive agents (robots or humans). The method relies on reciprocal velocity obstacles, given a global path, to compute a collision-free local path for a short duration. Turnwald \textit{et al.} \cite{Turnwald2019} presented a game-theoretic approach to SAN utilizing concepts from non-cooperative games and Nash equilibrium. This game theory-based SAN planner was evaluated against established planners such as reciprocal velocity obstacles or social forces, a variation of the Turing test was administered, which determines whether participants can differentiate between human motions and artificially generated motions. Aroor \textit{et al.} \cite{aroor2018online} formulated a Bayesian approach to develop an online global crowd model using a laser scanner. The model uses two new algorithms, CUSUM-A$^*$ (to track the spatiotemporal changes) and Risk-A$^*$ (to adjust for navigation cost due to interactions with humans), that rely on local observation to continuously update the crowd model. Unlike other model-based approaches, our method does not require any training data to perform socially-aware navigation.

Okal \textit{et al.} \cite{okal2016learning} presented a Bayesian Inverse Reinforcement Learning (BIRL) based approach to achieving socially normative robot navigation using expert demonstrations. The method extends BIRL to include a flexible graph-based representation to capture the relevant task structure that relies on collections of sampled trajectories. Kim \textit{et al.} \cite{kim2016socially} presented an Inverse Reinforcement Learning (IRL) based framework for socially adaptive path planning to generate human-like trajectories in dynamic human environments. The framework consists of three modules: a feature extractor, a learning module, and a path planning module. Kretzschmar \textit{et al.} \cite{kretzschmar2016socially} proposed a method to learn policies from demonstrations, to learn the model parameters of cooperative human navigation behavior that match the observed behavior concerning user-defined features. They used Hamiltonian Markov chain Monte Carlo sampling to compute the feature expectations. To adequately explore the space of trajectories, the method relied on the Voronoi graph of the environment from start to target position of the robot. 

Human motion prediction is vital in SAN as it allows the robot to plan and execute its motion behaviors according to the predicted human motion. In contrast to traditional human trajectory prediction approaches that use hand-crafted functions (social forces), Alahi \textit{et al.} \cite{alahi2016social} proposed an LSTM (Long Short-Term Memory) model which can observe general human motion and predict their future trajectories. Hamandi \textit{et al.} \cite{hamandi2018deepmotion} developed a novel approach using deep learning (LSTM) called DeepMoTIon, trained over public pedestrian surveillance data to predict human velocities. The DeepMoTIon method used a trained model to achieve human-aware navigation, where the robots imitate humans to navigate in crowded environments safely. 

Although SAN research is dominated by planning-related advancements, for a long-term HRI in human environments, we need to understand how cooperative human navigation works. Psychologists and roboticists are looking into social cues and their effects on HRI to better understand the socially-aware navigation problem. Suvei \textit{et al.} \cite{suvei2018would} investigated the problem of ``how a robot can get closer to people that it wants to interact with?" In a 2x2 between-subject study, the authors investigated the effect of social gaze cues on the personal space invasion using a human-sized mobile robot. The results from a 2 x 2 between-subjects experiment, with/without personal space invasion and with/without a social gaze cue, indicate that social gaze did play a role in participants' perceived safety of the robot. In another study, Tan \textit{et al.} showed that bystanders and observers of HRI felt safer around the robot than the actual interaction partner even though they both were in very close proximity to the robot~\cite{tan2018isociobot}. The authors justify the robot's design by collecting the responses of invited users evaluating the properties and appearance of the robot while interacting with it. Rajamohan \textit{et al.} \cite{rajamohan2019factors}, studied the role of robot height in HRI related to preferred interaction distance. Subjective data showed that participants regarded robots more favorably following their participation in the study. Moreover, participants rated the NAO most positively and the PR2 (Tall, with a fully expanded telescoping spine) most negatively.  

\subsection{Multi-Objective Optimization}
\begin{figure}[ht]
\centering
\includegraphics[height = 4.5cm]{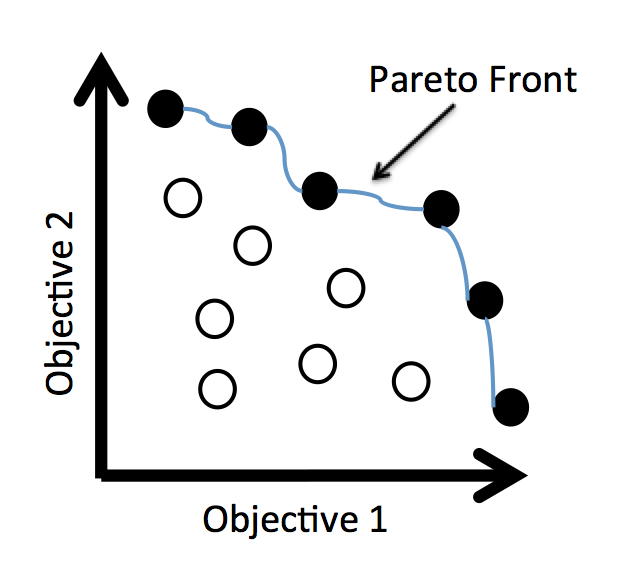}
\caption{\textit{Multi-Objective solution space} - The Pareto front contains the non-dominated solutions based on the the two objectives.}
\label{fig:Pareto_Front}
\end{figure}

It is easy to think of a task as a single objective function, where there is a goal or cost function that we are trying to either minimize or maximize. Ideally, this would always give the optimal solution for a task; however, this is not always the case. More often than not, multiple variables contribute to a cost function. An example of this is from basic economics, where there exists a market for a widget. As the supply of this widget goes up, the demand decreases and vice-versa. This phenomenon would be known as a supply and demand curve where one objective is the supply, and the other is the demand. In this case, the seller would want to find the optimal amount of supply, such that demand provides the optimal amount of profit. If graphed, this supply and demand curve, as shown in Figure~\ref{fig:Pareto_Front}, the points on the line would be Pareto optimal points; no point dominates any other. In this case, the seller is trying to maximize both objectives; therefore, the hollow circles are the dominated points as there exist solid circle points that are better in both objectives. Typically, multiple Pareto optimal points are forming a set, which is the solution type that many multi-objective algorithms use~\cite{coello1999comprehensive}.

Multi-objective optimization has already started to play a role in real-world applications~\cite{marler2004survey}. Some examples of real-world multi-objective scenarios are high speed civilian aircraft transportation~\cite{messac1996physical}, urban planning~\cite{balling1999multiobjective}, and for designing trusses~\cite{coello2000multiobjective}. The trajectory planner used in this paper builds off of a pre-existing one that utilizes a multi-objective approach along with linear combination~\cite{marder2010office}.

\subsubsection{PaCcET}
In some cases, optimizing a single objective does not yield the desired performance, and therefore multiple objectives need to be considered when evaluating a policy's fitness. A standard method is to multiply a preset scalar value to each objective's fitness score and then add them all together. In some domains, this can lead to an optimal set of policies; however, in some complicated domains, this method will yield sub-optimal policies. A solution to this is to use a multi-objective tool, such as PaCcET, to evaluate policies on multiple objectives~\cite{yliniemi2014paccet, yliniemi2015complete} properly. PaCcET works by first obtaining an understanding of the solution space and finding the Pareto optimal solutions. Next, PaCcET transforms the solution space and then compares each solution giving a single fitness score representative of how well each solution performed in the transformed space.

At a high level, PaCcET works by transforming the Pareto front in the objective space in a way that it is forced to be convex. Transforming to objective space allows the linear combination of transformed objectives to find a new Pareto optimal point. PaCcET iteratively updates this transformation to force non-explored areas of the Pareto front to be more highly valued than points dominated by the Pareto front or points that are on the explored areas of the Pareto front.

PaCcET has seen a variety of applications: it has been used to extend the life of a fuel cell in a hybrid turbine-fuel cell power generation system~\cite{colby2016multiobjective}, the operation of the electrical grid on naval vessels~\cite{Electric_Ship_Technologies_Symposium}, the coordination of multi-robot systems~\cite{Yliniemi2014considerations}, and for the efficient operation of a distributed electrical microgrid~\cite{sarfi2017novel}, where a series of small power generation systems coordinate to meet the demands of consumers. In each of these applications, it has been shown that PaCcET functions at or above the solution quality of other techniques like NSGA-II or SPEA2~\cite{yliniemi2014paccet}, with as low as one-tenth of the run-time.

\begin{figure}[ht]
\centering
\includegraphics[height = 4.5cm]{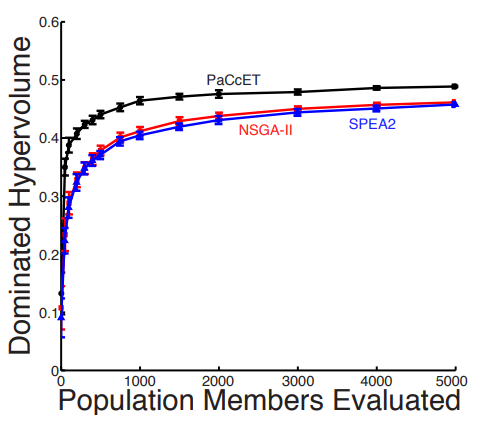}
\caption{\textit{PaCcET computational speed} - Percentage of hypervolume dominated in Kursawe's (KUR) problem in comparison with two successful multi-objective methods, SPEA2 and NSGA-II. This plot shows that PaCcET proceeds faster towards the Pareto front.}
\label{fig:PaCcET_speed}
\end{figure}

For this project, PaCcET was used over other multi-objective tools because of its computational speed~\cite{yliniemi2014paccet}, as shown in Figure~\ref{fig:PaCcET_speed}. PaCcET was used to evaluate the possible trajectories developed in the local planner. At each time step, the sensor data is analyzed, and the desired features are evaluated for each of the potential future trajectory points. PaCcET then uses the fitness values for each feature of every future trajectory to develop the solution space and obtain the optimal future trajectory. Since at each time step, future trajectories are developed independently, PaCcET develops a brand new solution space at each time step. By developing a new solution space at each time step, the local planner can be optimized in real-time.

\section{Method}
\label{sec:method}
\begin{figure}[ht]
\centering
\includegraphics[height = 4.5cm]{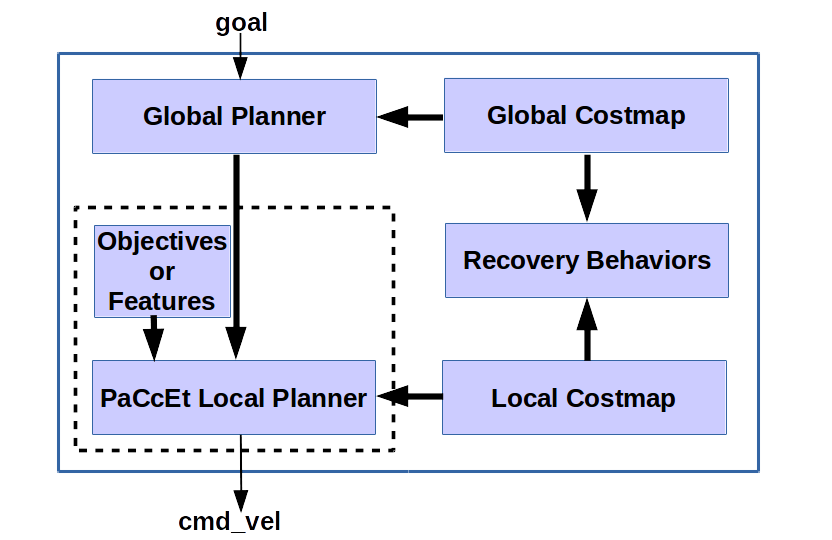}
\caption{Block diagram of the proposed approach, a modification of ROS navigation stack's local planner using PaCcET-based non-linear optimization.}
\label{fig:blockdiagram}
\end{figure}

\begin{figure}[ht]
\centering
\includegraphics[height = 4.5cm]{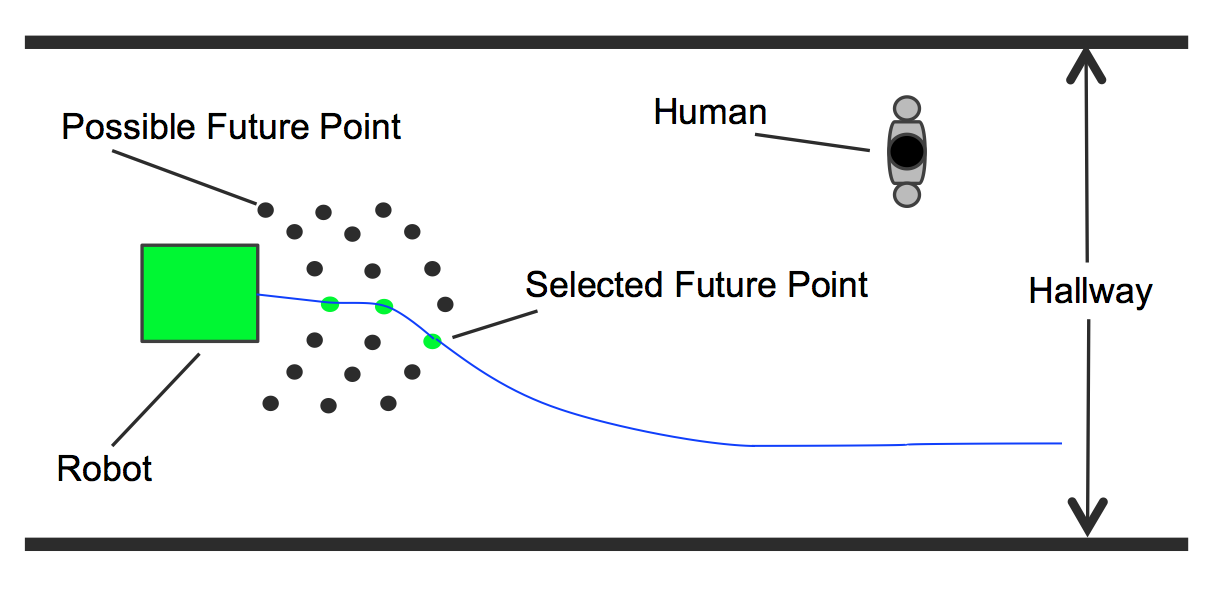}
\caption{\textit{Navigation Planner} - The navigation planner selects a short-term trajectory (green points represent potential trajectory end-points) from the pool of possible trajectories (black points), optimized for adherence to a long-term plan (blue line), obstacle avoidance, and progress toward a goal, and in the case of this paper, interpersonal distance.}
\label{fig:Nav_Stack}
\end{figure}

In this section, we detail our methodology of a socially-aware navigation planner, the features or objectives that we used to optimize the trajectories, and how PaCcET was implemented in the local trajectory selection process. Figure \ref{fig:blockdiagram} shows the overall high-level block diagram of the proposed approach. It is built on top of the well-established ROS navigation stack by modifying the local planner using PaCcET.  The overall function of the local trajectory planner at each time step is to generate an array of possible future trajectory points and evaluate each future trajectory point based on a predefined feature set as shown in Figure~\ref{fig:Nav_Stack}. In previous work, the features were assumed to have either no relationship or a simple linear relationship with one another; however, this is not always the case. Therefore, we need to consider the possibility that the features are dependent on each other and have nonlinear relationships.

\subsection{Features/Objectives}
In the traditional navigation planner, the features extracted were each assigned their own cost (e.g., the path distance cost ($\Delta_{path}$), the length that the robot has already traveled, the goal distance cost ($\Delta_{goal}$), the distance the robot is from the goal)~\cite{marder2010office}. The path distance will have a linear relationship with the goal distance since the change in one has a direct linear impact on the other. Once each feature has a cost associated with it, each cost is multiplied by a pre-tuned scalar and then added together, thus giving a linear combination, or weighted sum, in this case, the cost function shown in Equation~\ref{eq:Cost function example}. We can think of this cost function as an objective, where each possible future trajectory point has a cost or fitness associated with that objective. Since the purpose is to minimize the overall cost function, the planner will take the best path possible that minimizes the function, which in this case, will minimize both features.

\begin{align}
\label{eq:Cost function example}
\text{cost}({v}_{x},{v}_{y},{v}_{\theta}) = \alpha(\Delta_{path})+\beta(\Delta_{goal})
\end{align}

This cost function has been adapted to include a heading difference ($\Delta_{heading}$) feature and an occupancy ($\Delta_{occ}$) cost feature, where the heading difference is the distance that the robot is from the global path and the occupancy cost is the cost used to keep the robot from hitting something. The same approach as in the previous cost function is taken in Equation~\ref{eq:Cost function}. By taking a closer look at just the heading difference and how that might affect the path distance or the goal distance, it becomes less clear if there is only a linear relationship between the four. For example, if there is an obstacle in the robot's path, it will try and minimize goal distance by changing its heading, thus increasing the heading feature cost. In turn, this also increases the path distance cost, though this may or may not be linear.

\begin{equation}
\begin{split}
\label{eq:Cost function}
\text{cost}({v}_{x},{v}_{y},{v}_{\theta}) &= \alpha(\Delta_{path})+\beta(\Delta_{goal})+\gamma(\Delta_{heading})\\
& +\delta(\Delta_{occ})
\end{split}
\end{equation}

Building upon prior work done in this area, we include socially-aware navigation features such as interpersonal distance ($ID$), distance from a group ($GD$), and distance from a social goal ($SGD$). As a way to dissuade the robot from getting too close to a human, a cost function was developed to penalize the robot at an exponential rate as the interpersonal distance decreases, as seen in Equation~\ref{eq:Interpersonal Distance} (for every human in the interaction scenario). Although we could penalize the robot based on this at all times, it is not necessary if the interpersonal distance is so significant that it would not be considered as a socially inappropriate distance. Therefore the robot is only penalized if the interpersonal distance is less than or equal to 1.5 meters. The interpersonal distance threshold was chosen to be 1.5 meters to ensure that the robot remains in social space and does not invade the personal space of the person~\cite{hall1966hidden}.

\begin{equation}
\label{eq:Interpersonal Distance}
\text{$ID_f$} = e^{1/ID}
\end{equation}

For the robot to not get too close to a group of people or not get in between them, we penalized the robot based on $GD$ whenever it is close to a group using Equation~\ref{eq:Group Distance}.
\begin{equation}
\label{eq:Group Distance}
\text{$GD_f$} = e^{1/GD}
\end{equation}

Contrary to Equation~\ref{eq:Interpersonal Distance} and~\ref{eq:Group Distance}, Equation~\ref{eq:Social Goal Distance} is akin to a reward to get closer to a social goal rather than an actual goal. With this feature in place, the robot tends to reach a social goal for a particular scenario while still adhering to the final goal location. For example, the social goal for reaching the front of a desk when others are waiting in a line is the end of the line. So, the robot will reach the social goal first (end of the line) and eventually reaches the desk when it is the robot's turn.  

\begin{equation}
\label{eq:Social Goal Distance}
\text{$SGD_f$} = e^{SGD}
\end{equation}

Instead of adding these features cost into the previous cost function, as in Equation~\ref{eq:Cost function}, we assume that its relationship with other features might be nonlinear and therefore gets treated as separate objectives. Since we know that the above cost function, Equation~\ref{eq:Cost function}, works sufficiently enough from previous work \cite{marder2010office}, we can treat it as its objective as well. Instead of optimizing just one objective, we need to optimize multiple objectives, hence our multi-objective approach. Using a multi-objective tool like PaCcET requires computational time, and since this is intended to work in real-time, any chance to improve the computation time should be utilized. Treating the first four features used in the previous cost equation as a single objective not only speeds up this process but, in turn, allows for the possibility to add even more features to our local trajectory planner. Using PaCcET to do the multi-objective transformations, we essentially get a new cost function with a PaCcET fitness denoted by $P_f$, which was modeled under the assumption of nonlinear relationships between the objectives. Equation~\ref{eq:PaCcET fitness example} shows how $P_f$ is a transformation function dependent on multiple variables.

\begin{align}
\label{eq:PaCcET fitness example}
\text{$P_f$}= T_f(Obj_1,Obj_2,....,Obj_n)
\end{align}

In this work, we are only interested in objectives like interpersonal distance, distance from a group, and distance from the social goal. The first objective is the original cost function (Equation~\ref{eq:Cost function}), which is the linear combination of the path distance, goal distance, heading difference, and occupancy cost. The remaining objectives are the social features, such as:interpersonal distance, distance from social goal, etc. Equation~\ref{eq:PaCcET fitness} shows the PaCcET fitness function with our proposed objectives.

\begin{align}
\label{eq:PaCcET fitness}
\text{$P_f$}= T_f({cost}({v}_{x},{v}_{y},{v}_{\theta}),ID_{f1}, .. , ID_{fn}, GD_f, SGD_f)
\end{align}

where, $ID_{f1}$, .. $ID_{fn}$ are the cost functions associated with interpersonal distance between $n$ people and the robot.

\subsection{Trajectory Planning}
The robot's trajectory planning algorithm can be broken into three parts, the global planner, the local planner, and low-level collision detection and avoidance. The global trajectory planner works by using knowledge of the map to produce an optimal route given the robot's starting position and the goal position. The global path is created as a high-level planning task based upon the robot's existing map of the environment; this is regenerated every few seconds to take advantage of shorter paths that might be found or to navigate around unplanned obstacles. The traditional local planner's role is to stay in line with the global path unless an obstacle makes it deviate from the global path. The low-level collision detector works by stopping the robot if it gets too close to an object. We use the traditional global trajectory planner and low-level collision detector~\cite{marder2010office} and make adaptations to the local trajectory planner to incorporate interpersonal distance and PaCcET.

The algorithm \ref{algo:Local Trajectory Planner}, algorithm \ref{algo:Generate Trajectory} and algorithm \ref{algo:PaCcET} can be summarized as follows:

\begin{enumerate}
    \item Discretely sample the robot control space.
    \item For each sampled velocity ($V_x$, $V_y$, and $V_\theta$), perform a forward simulation from the robot's current state for a short duration to see what would happen if the sampled velocities were applied. This is robot-specific, based on the footprint of the robot.
    \item Score the trajectories based on metrics.
    \begin{enumerate}
        \item Score each trajectory from the previous step for metrics like distance to obstacles, distance to a goal, etc. Discard all the trajectories that lead to a collision in the environment. 
        \item For each of the valid trajectories, calculate the social objective fitness scores like interpersonal distance and other social features and store all the valid trajectories.
    \end{enumerate}
    \item Perform Pareto Concavity Elimination Transformation (PaCcET) on the stored trajectories to get a PaCcET fitness score and sort the trajectories from lowest to highest PaCcET fitness score.
    \item For a given time step, grab the trajectory with index 0 from the sorted list of valid trajectories.   
\end{enumerate}

Algorithm~\ref{algo:Local Trajectory Planner} shows the primary functions of the local trajectory planner and how the future trajectory points were stored to be used with PaCcET. The trajectory planner is called every time step, which in this case, is every $0.1$ second. Once the trajectory planner is called, the \texttt{Transform\_Human\_State} function is called to transform human pose to the robot's odometry reference frame, which allows the interpersonal distance corresponding to each possible trajectory to be calculated in the \texttt{Generate\_Trajectory} function. There are two methods of calculating the possible trajectories. The first is assuming that the robot can only move forward, backward, and turn. To produce the possible trajectories for this physical set up we loop through every combination of a sample of linear velocities ($V_x$) and angular velocities ($V_\theta$) to generate trajectories (For a holonomic robot, a slight change in $V_y$ is also used to generate possible trajectories). Once a trajectory is created, we determine if it is valid based on the constraints for the first objective. For example, trajectories that would make the robot hit a wall, obstacle, or human are not considered strong trajectories and, therefore, will not be stored in the \texttt{Store\_Trajectory} function. By not storing these invalid trajectories, the speed at which PaCcET runs can be improved.

\begin{algorithm}[ht]
\DontPrintSemicolon
\KwIn{$V_x$ samples, $V_\theta$ samples, $H_s$, $R_s$}
\KwOut{\text{$Best\_Trajectory$} $(\mathcal{T_B})$}
\caption{\texttt{Local Trajectory Planner Algorithm}. The trajectory planner generates multiple trajectories ($T$) given a number of $V_x$ samples and $V_\theta$ samples and calculates the independent cost for each feature. The cost for each feature is based on the robots sensing of the human's state ($H_s$) and the robot's state ($R_s$). At the end of a time step, the best trajectory $(\mathcal{T_B})$ out of all valid trajectories $(\mathbb{T})$ is returned.}
\label{algo:Local Trajectory Planner}
\For{Each time step}
{
    \texttt{$Transform\_Human\_State$($H_s$,$R_s$)}\;
    \For{Each $V_x$}
    {
        $\mathcal{T} \leftarrow \texttt{$Generate\_Trajectory$($T$, $H_s$)}$\;
        \If{valid trajectory}
        {
            \texttt{$Store\_Trajectory$($\mathcal{T}$)}\;
        }
        \For{Each $V_\theta$}
        {
            $\mathcal{T} \leftarrow \texttt{$Generate\_Trajectory$($T$, $H_s$)}$\;
            \If{Valid Trajectory}
            {
                \texttt{$Store\_Trajectory$($\mathcal{T}$)}\;
            }
        }
    }
    \If{Holonomic Robot}
    {
        $\mathcal{T} \leftarrow \texttt{$Generate\_Trajectory$($T$, $H_s$)}$\;
        \If{Valid Trajectory}
        {
            \texttt{$Store\_Trajectory$($\mathcal{T}$)}\;
        }
    }
    \texttt{$Run\_PaCcET$($\mathbb{T}$)}\;
    $Return\ \mathcal{T_B}$\;
}
\end{algorithm}

The second method assumes that the robot is capable of holonomic movement and can translate with any $V_x,V_y,V_\theta$ that are less than velocity limits. Given these movements, we again loop through all the possible movements given the predefined number of $V_x$ samples, $V_y$ samples, and $V_\theta$ samples. Again, if the trajectories are valid, they are stored. Once all the valid trajectories are stored for all possible movements, the \texttt{Run\_PaCcET} function runs, giving back the best possible trajectory, $(\mathcal{T_B})$, based on its multi-objective transformation process.

In order to run a multi-objective tool like PaCcET, each objective's fitness needs to be calculated. Algorithm~\ref{algo:Generate Trajectory} details the \texttt{Generate\_Trajectory} function from Algorithm~\ref{algo:Local Trajectory Planner}. The first function that needs to be performed is the \texttt{Calculate\_State} function as the robot's position and velocity are used to determine the fitness values for the objectives. Using the state information the \texttt{Compute\_Path\_Dist}, \texttt{Compute\_Goal\_Dist}, \texttt{Compute\_Occ\_Cost}, and \texttt{Compute\_Heading\_Diff} functions are used to calculate the fitness values associated with the four pieces of the first objective. Using those fitness values, the first objective's fitness is calculated by the \texttt{Compute\_Cost} function. Distance-based features like interpersonal distance of each person, group distance, and social goal distance are calculated, as shown in Algorithm~\ref{algo:Generate Trajectory} lines~\ref{line7} - \ref{line11}. Once all the objectives have their fitness values, the trajectory along with the fitness values is returned to the local trajectory planner algorithm, which saves all the valid trajectories and calls PaCcET Algorithm~\ref{algo:PaCcET} to output socially appropriate trajectories.

\begin{algorithm}[ht]
\DontPrintSemicolon
\KwIn{$T$, $H_s$}
\KwOut{\text{$T$}}
\caption{\texttt{Generate Trajectory Algorithm}. The generate trajectory function takes in an instance of a trajectory ($T$) and the humans' state ($H_s$) to compute the cost function for each feature. The trajectory ($T$) is then returned to the local trajectory planner.}
\label{algo:Generate Trajectory}
$\mathcal{S} \leftarrow \texttt{$Calculate\_State$($\mathcal{T}$)}$\;
$\texttt{$path\_dist$} \leftarrow \texttt{$Compute\_Path\_Dist$($\mathcal{S}$)}$\;
$\texttt{$goal\_dist$} \leftarrow \texttt{$Compute\_Goal\_Dist$($\mathcal{S}$)}$\;
$\texttt{$occ\_cost$} \leftarrow \texttt{$Compute\_Occ\_Cost$($\mathcal{S}$)}$\;
$\texttt{$heading\_diff$} \leftarrow \texttt{$Compute\_Heading\_Diff$($\mathcal{S}$)}$\;
$\texttt{$cost$} \leftarrow \texttt{$Compute\_Cost$($path\_dist$, $goal\_dist$,} \linebreak \texttt{$occ\_cost$, $heading\_diff$)}$\;
\For{Each person}
{
$\mathcal{ID} \leftarrow \texttt{$Calculate\_Interpersonal\_Distance$($H_s$, $\mathcal{S}$)}$\; \label{line7}
}
$\mathcal{GD} \leftarrow \texttt{$Calculate\_Group\_Distance$($H_s$, $\mathcal{S}$)}$\;
$\mathcal{SGD} \leftarrow \texttt{$Calculate\_Social\_Goal\_Distance$($H_s$, $\mathcal{S}$)}$\; \label{line11}

$Return\ {\text{$Trajectory$($\mathcal{T}$)}}$\;
\end{algorithm}

\subsection{Integrating PaCcET}
At the end of Algorithm~\ref{algo:Local Trajectory Planner}, all the valid trajectories have been stored along with their objective fitness scores in a vector of type trajectory. Algorithm~\ref{algo:PaCcET} details the primary functions for determining a single fitness value from multiple objectives. In order to run PaCcET, the objectives for each trajectory must be stored in a vector of type double, which is done in the \texttt{Store\_Objectives} function. Before running PaCcET's primary functions, an instance of PaCcET must be created. Next, the solution space and Pareto front are created by giving each trajectory to the \texttt{Pareto\_Check} function. Now that the Pareto front and its geometry have been calculated, PaCcET can transform the solution space and give a single fitness value for each trajectory in the \texttt{Compute\_PaCcET\_Fitness} function. Once each trajectory has its PaCcET fitness, they are sorted from best to worst in the \texttt{Sort\_Trajectories} function, which allows the function to not only ascertain the best trajectory easily, but is also useful for debugging purposes. Algorithm~\ref{algo:PaCcET} concludes by returning the best trajectory to the local trajectory planner algorithm.

\begin{algorithm}[ht]
\DontPrintSemicolon
\KwIn{$\mathbb{T}$}
\KwOut{\text{$\mathcal{T_B}$}}
\caption{\texttt{PaCcET Alogrithm}. PaCcET ($P$),takes in the vector of valid possible trajectories $\mathbb{T}$ to compute the multi-objective space and the PaCcET fitness ($P_f$) for each trajectory.}
\label{algo:PaCcET}
\For{Each trajectory}
{
    \texttt{$Store\_Objectives$($\mathcal{T}$)}
}
$\mathcal{P} \leftarrow \texttt{$Initialize\_PaCcET$()}$\;
\For{Each trajectory}
{
    \texttt{$Pareto\_Check$($\mathcal{T}$)}\;
}
\For{Each trajectory}
{
    $\texttt{$P_f$} \leftarrow\texttt{$Compute\_PaCcET\_Fitness$($\mathcal{T}$)}$\;
}
\texttt{$Sort\_Trajectories$($\mathbb{T}$)}\;
$Return\ \mathcal{T_B}$\;
\end{algorithm}

\subsection{Social Goal}

\begin{figure}[ht]
\centering
\includegraphics[height = 4.5cm]{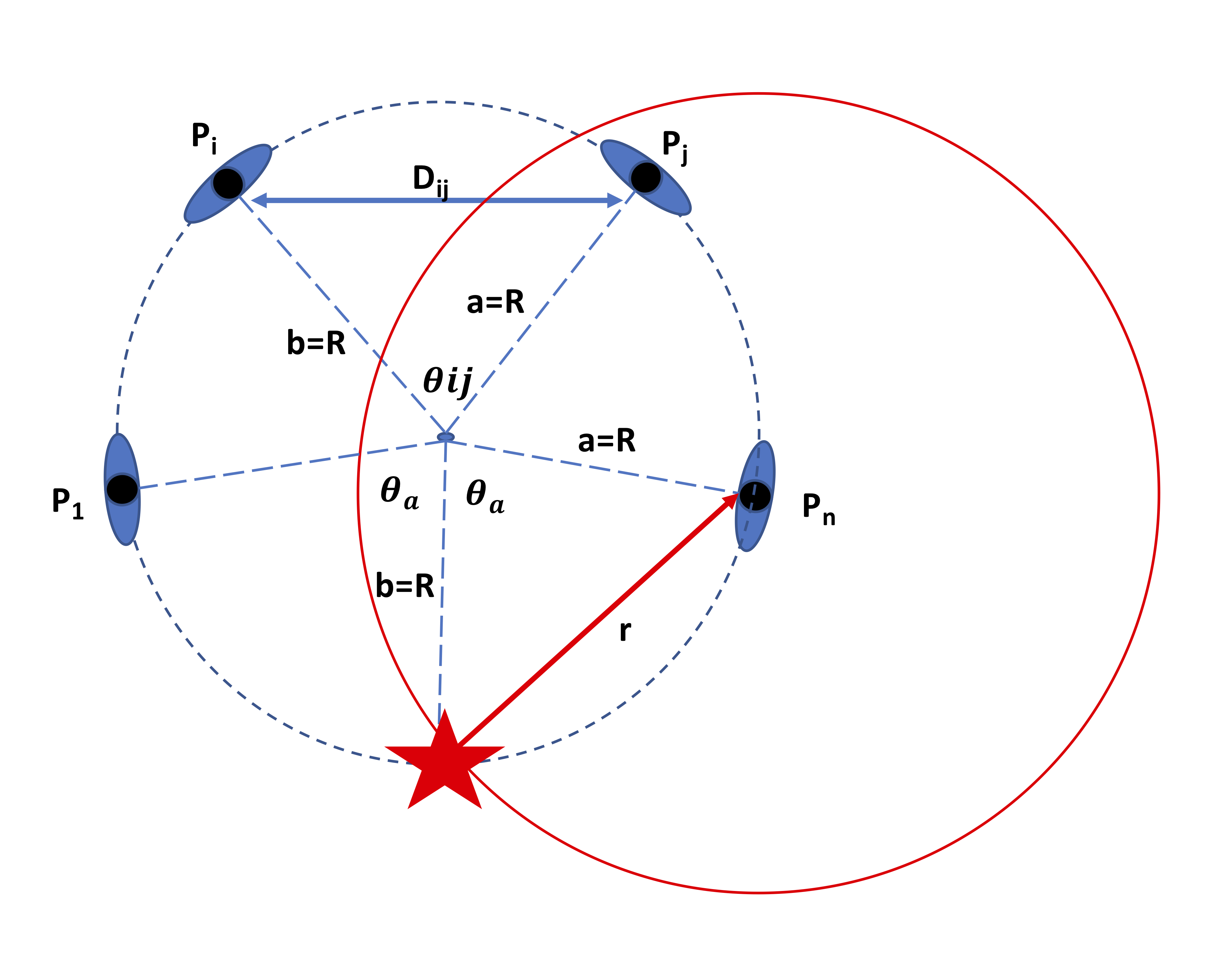}
\caption{Figure illustrating the computation of social goal in the O-formation scenario. The red star represents a goal that respects social norms.}
\label{fig:circle_formation}
\end{figure}

Computing the social goal location is important because often, the actual goal location may not be an appropriate location for interaction, and explicitly commanding the social goal would not be possible. A social goal can be defined as an appropriate location for a robot to involve in human-robot interaction. For example, in a front desk-like scenario, the end of the line can be considered social. For this work, the social goals for each interaction scenario were geometrically computed, for \textit{O-formation scenario (joining a group)}, we fit a circle with the people in a group and find a socially appropriate spot to join the group as shown in Figure~\ref{fig:circle_formation}.

We found angle made by every person with the center of the formed circle using the law of cosines, equation~\ref{eq:law of cosines} as shown below:

\begin{equation}
\label{eq:law of cosines}
c^2 = a^2+b^2-2ab * cos(\theta)
\end{equation}
\begin{equation}
    \theta_{ij} = cos^{-1}[(2R^2-D_{ij})/2R^2]
\end{equation}

Where $D_{ij}$ is the Euclidean distance between person $i$, $j$, and $R$ is the radius of the circle formed by all the people in the group. Out of all the $\theta_{ij}$'s, we pick one half of the widest angle as approach angle denoted by $\theta_a$. Now, joining a group problem (O-formation) boils doing to finding the intersection of two circles, one formed by the people in the group and the other formed in the wide-open sector with the center as the locations of either of the people making the widest sector. The equations of the two circles to solve for are given as follows:

\begin{figure}[ht]
\centering
\includegraphics[height = 4.5cm]{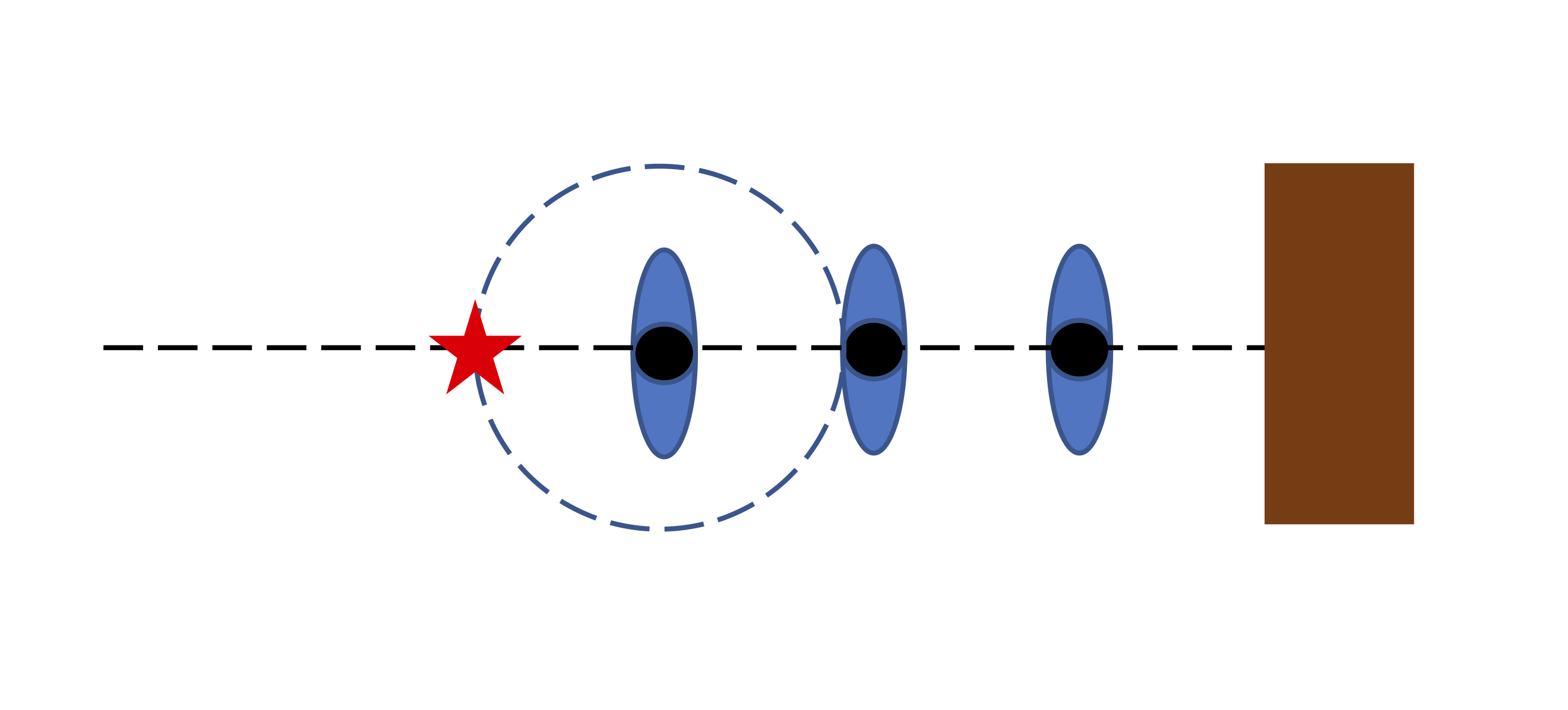}
\caption{Figure illustrating the computation of social goal in waiting in a queue scenario. The red star represents the social goal.}
\label{fig:line_formation}
\end{figure}

\begin{equation}
\label{eq:first_circle}
    (x-h)^2+(y-k)^2 = R^2
\end{equation}
\begin{equation}
\label{eq:second_circle}
    (x-h_p)^2+(y-k_p)^2 = r^2
\end{equation}

Where, $(h, k)$ is the center of the circle formed by the group of people, $(h_p, k_p)$ is the location of one of the person that formed the widest sector. The radius $r$ in Equation \ref{eq:second_circle} is obtained by solving for $c$ in Equation \ref{eq:law of cosines} where, $\theta = \theta_a$, $a$ and $b$ equals $R$, radius of the group formation. There are two solutions when solving Equations \ref{eq:first_circle} and \ref{eq:second_circle}, we further filter one social goal from the two solutions. 

Similarly, we can fit a straight line as shown in Figure~\ref{fig:line_formation} for \textit{waiting in a queue scenario}, and social goal location would be the end of the line considering the personal space of the last person in the line. Hence, in this case, the solution boils down to solving for the intersection of a line and a circle.
The equation of the line formed by the people can be found by fitting a line of form $y = mx + c$ with the people's locations. The circle formed using the last person's location as the center and a comfortable distance that the robot should maintain around the last person as the radius is of the form $(x-k)^2+(y-h)^2=r^2$. The two solutions to the line and circle intersection can be obtained using quadratic roots, and the social goal is further filtered to the solution farthest to the actual goal (desk). 

The social goal calculation for Scenario 2 (\textit{art gallery}) is hand-selected. Computing the social goal in this scenario is beyond the scope of this paper. The art gallery scenario requires a perception pipeline that can detect the location of the art on display, the area, perimeter of the artwork to compute a useful social goal location to interact with a person viewing the art or to determine the activity zone to avoid traversing it. The social goal for Scenario 2 will be addressed in our on-going work on USAN, see Section~\ref{sec:discussion}.

In this section, we presented an in-depth illustration of our proposed PaCcET-based social planner, various cardinal objectives related to SAN, methods to determine social goals using spatial information such as locations of people. The next section, Section~\ref{sec:results}, presents the results of robots performing appropriate navigation behaviors in simulation and real-world environments.

\section{Results}
\label{sec:results}

\begin{figure*}[th]
\centering
\begin{subfigure}{0.47\textwidth}
\centering
\includegraphics[height=4.5cm]{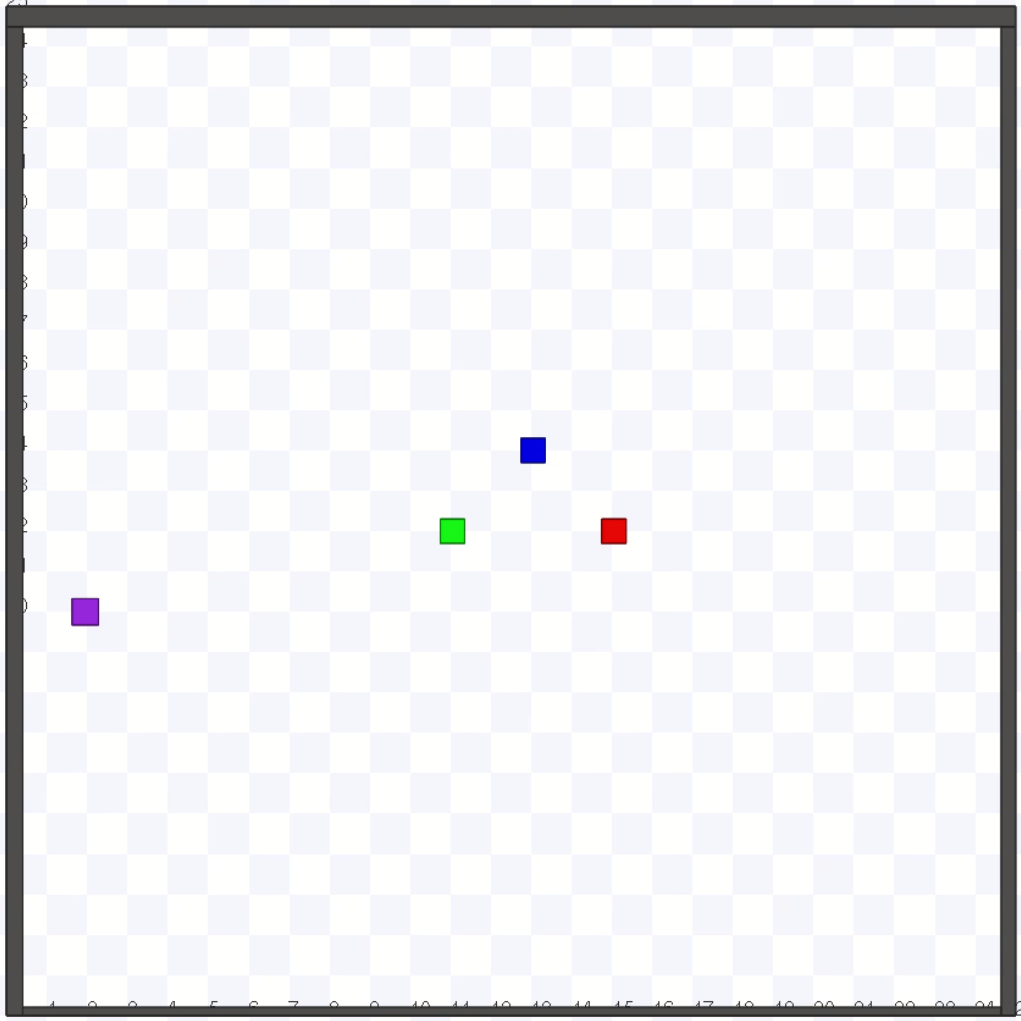} 
\caption{Stage, a 2D simulator.}
\label{fig:stage}
\end{subfigure}
\hspace{4mm}
\begin{subfigure}{0.47\textwidth}
\centering
\includegraphics[height=4.5cm]{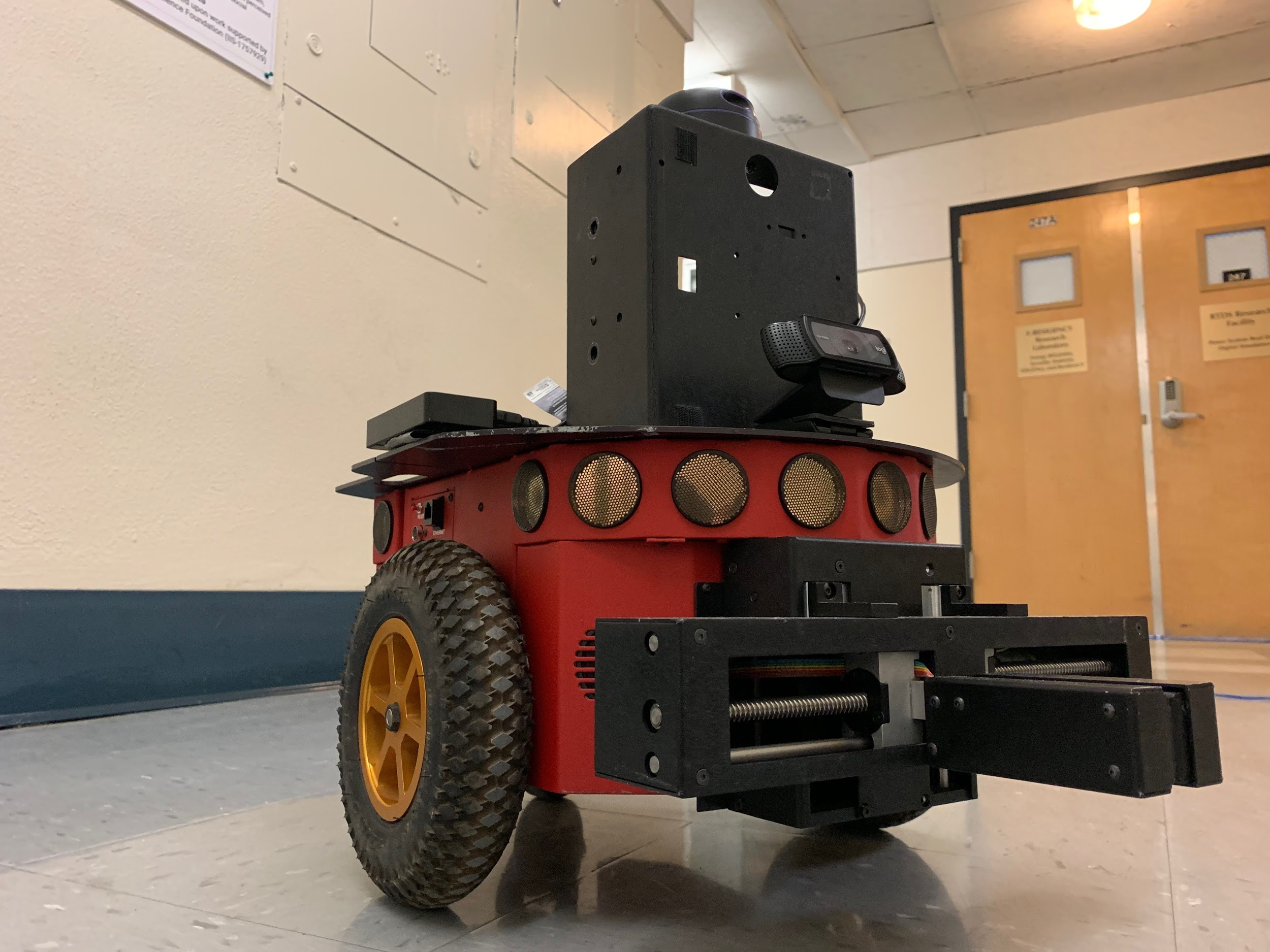}
\caption{Pioneer 3DX robot.}
\label{fig:robot}
\end{subfigure}
\caption{Platforms used to validate our proposed social planner}
\label{fig:platforms}
\end{figure*}

To validate our proposed approach, we considered four different scenarios, namely, \textit{a hallway scenario, an art gallery scenario, forming a group, and waiting in a queue}. All this was tested in simulation using the 2D simulator, Stage~\cite{gerkey2003player} on a machine with an Intel 6th-generation i7 processor @3.4 GHz, 32 GB of RAM. The simulated environment for each hallway experiment was the second-floor hallway of the Scrugham Engineering and Mines building at the University of Nevada, Reno. The map of the building used in the simulation was built using the gmapping package for SLAM on the PR2. The simulated PR2 is comparable to the real-world PR2 for sensing and movement capabilities and is using AMCL for localization on the map. The simulated PR2 uses a 30-meter range laser scanner that is identical to the real PR2 robot's laser scanner's capability and the humans in the simulation exhibit very simple motion behaviors. Follow-up scenarios are simulated in a 25m x 25m open space in the stage environment, as shown in Figure~\ref{fig:stage}. PR2 robot was simulated to run both traditional planner and our modified PaCcET based planner. In Figure~\ref{fig:stage}, the purple agent is the simulated PR2, and the rest of the agents are humans formed as a group. For real-world validation, we used an upgraded Pioneer 3DX platform, shown in Figure~\ref{fig:robot}. The Pioneer robot that we used is equipped with an RPLIDAR-A3, a 30-meter range laser scanner with a 360\textdegree field of view, and a webcam as sensors for perception. For detecting people using a laser scanner, we used the work of Leigh \textit{et al.}~\cite{leigh2015person}. The robot's computational unit is also upgraded to a laptop with an Intel Core i7-7700HQ CPU @ 2.80 GHz x 8 processors, 16 GB RAM GeForce GTX 1050 Ti GPU with 4GB of memory. The pioneer robot also uses AMCL for localization on the map. The hallway scenarios on the real robot were validated in the same location as the simulation experiments. \textit{Art gallery, waiting in line, and group formation scenarios} are validated in the lobby area (7m x 7m approx.) situated on the first floor of the Scrugham Engineering and Mines building of University of Nevada, Reno.


\subsection{Simple scenario}
\label{sec:simple_results}
\begin{figure*}[th]
 \centering
\begin{subfigure}{0.47\textwidth}
\centering
\includegraphics[height=4.5cm]{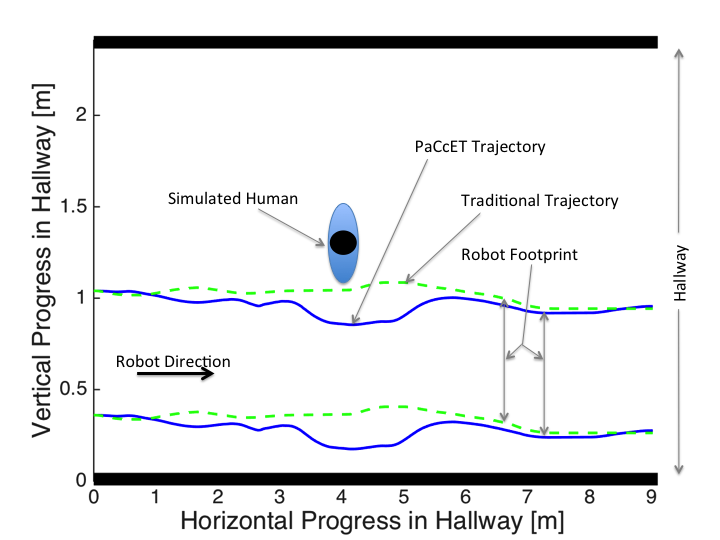} 
\caption{Scenario 1: Simulated PR2 passing a simulated stationary human in a narrow hallway.}
\label{fig:subim1}
\end{subfigure}
\hspace{4mm}
\begin{subfigure}{0.47\textwidth}
\includegraphics[height=4.5cm]{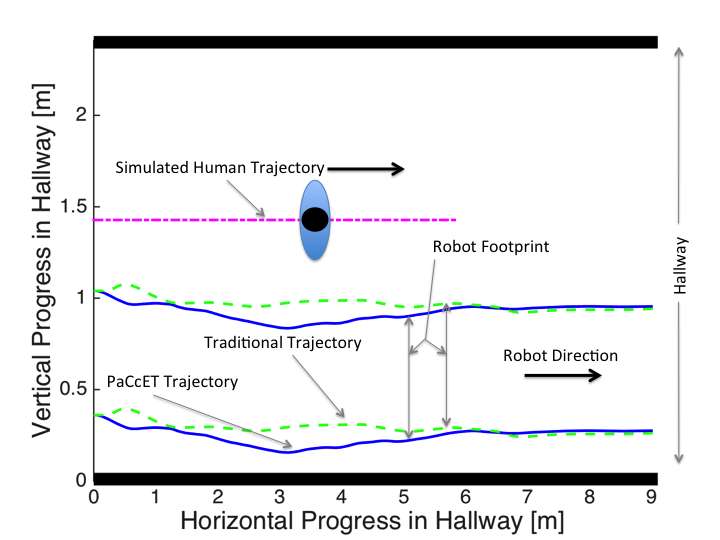}
\caption{Scenario 1: Simulated PR2 passing a simulated human walking in the same direction as the PR2 in a narrow hallway.}
\label{fig:subim2}
\end{subfigure}
\begin{subfigure}{0.47\textwidth}
\includegraphics[height=4.5cm]{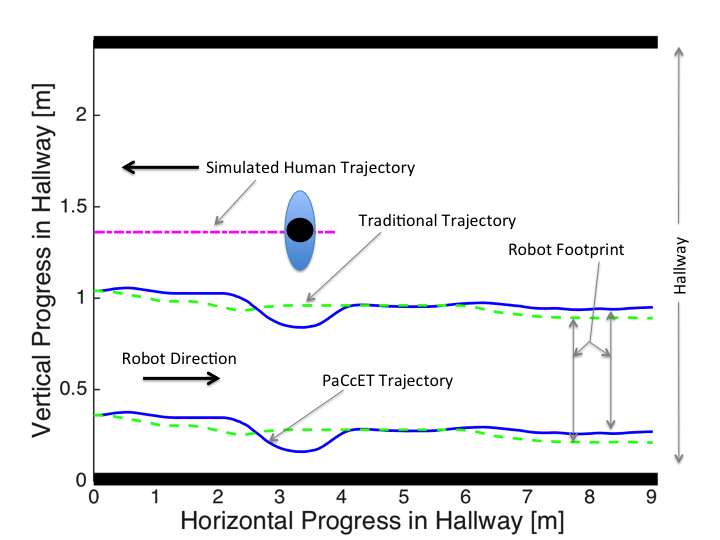}
\caption{Scenario 1: Simulated PR2 encounters a simulated human passing on the appropriate side of a narrow hallway.}
\label{fig:subim3}
\end{subfigure}
\hspace{4mm}
\begin{subfigure}{0.47\textwidth}
\includegraphics[height=4.5cm]{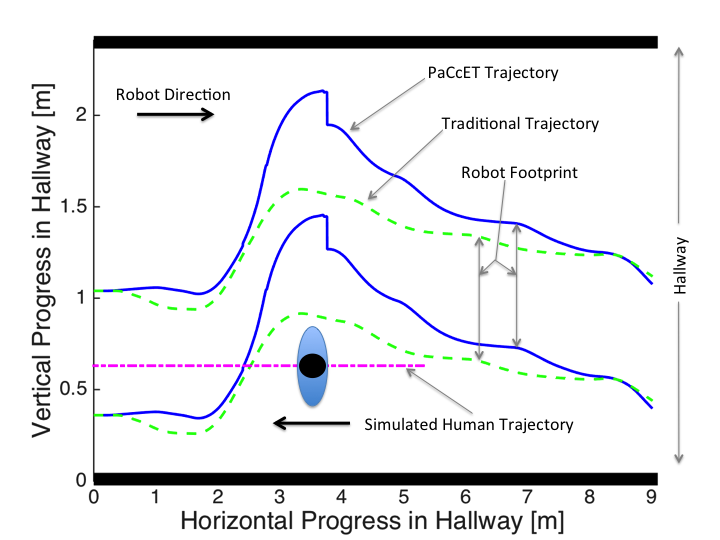} 
\caption{Scenario 1: Simulated PR2 encounters a simulated human walking on the inappropriate side of a narrow hallway in opposite direction.}
\label{fig:subim4}
\end{subfigure}
\caption{Simple two-objective optimization scenarios with a single simulated human. The simulated human trajectory is shown using a dotted magenta line, trajectory of traditional planner is represented using a dotted green lines (two lines to represent footprint of the simulated PR2) and PaCcET based SAN trajectory is represented using a solid blue line (two lines to represent footprint of the simulated PR2). Direction of simulated human and PR2 are represented using arrows.}
\label{fig:prior_results}
\end{figure*}

The hallway scenario was divided into four sub-scenarios, namely, passing a stationary human, passing a walking human in the same direction, an encounter with a human walking on the appropriate side, and passing a walking human in the opposite direction. 

In the first experiment, the simulated robot was tasked with getting to a goal while passing close to a static simulated human. Figure~\ref{fig:subim1} shows that when using the traditional planner, the robot made sure to avoid a collision with the simulated human but did not consider any social distance. The same will be the case for the other experiments as well since the traditional planner does not consider interpersonal distance into its cost function. The PaCcET-based planner did consider interpersonal distance, and therefore the robot deviated from a more straight-lined path as a way to satisfy the second objective. Once the threshold for the interpersonal distance was no longer an issue, the robot only needed to minimize the first objective; therefore, returning to a straight-line path. It is worth noting that in all the conducted experiments, the robot also considered a wall as an obstacle and was required to disregard trajectories that would lead to a collision, which is why the robot refrained from deviating any further from the global trajectory.

The second experiment was developed to mimic a passing scenario where the robot has a set goal but needs to pass by a simulated human who is traveling much slower in the same direction. Figure~\ref{fig:subim2} shows that the traditional trajectory planner merely made sure that a collision would not take place as it tried to minimize its cost function. The PaCcET-based planner deviated from its global trajectory in order to consider the interpersonal distance objective, then returned to the global trajectory once the threshold for the interpersonal distance was no longer an issue.

Similar to the previous experiment, the third experiment involves both the simulated human and robot moving; however, in this case, the simulated human is now moving at a normal walking speed in the opposite direction of the robot—the robot and simulated human pass close to one another but not close enough to cause a collision. Figure~\ref{fig:subim3} shows that the traditional trajectory planner altered its path ever so slightly to ensure that a collision would not happen, where the PaCcET-based trajectory planner not only ensured that a collision would not take place but also considered interpersonal distance and provided the simulated human with additional space while passing.

The previous experiments show that when using a {PaCcET}-based trajectory planner, interpersonal distance can be considered when selecting a local trajectory in both static and dynamic conditions where a collision is not imminent. However, the case of a collision that would occur unless either the simulated human or the robot moves out of the way also needs to be considered. This experiment considers a simulated human who is not paying attention or unwilling to change their course and walking directly towards the robot. Figure~\ref{fig:subim4} shows that the traditional trajectory planner was successful at avoiding the collision as expected; however, it did so while minimizing its cost function as much as possible, which caused the robot to get very close to the simulated human. When using the PaCcET-based trajectory planner, the robot not only avoided the collision but also gave the simulated human additional space to satisfy the interpersonal distance objective. It is worth noting that once the interpersonal distance threshold was no longer an issue, the robot used its holonomic movement for a short time as a way to quickly minimize the heading difference portion of the original cost function objective.

\begin{figure}
\centering
\includegraphics[width=0.9\textwidth]{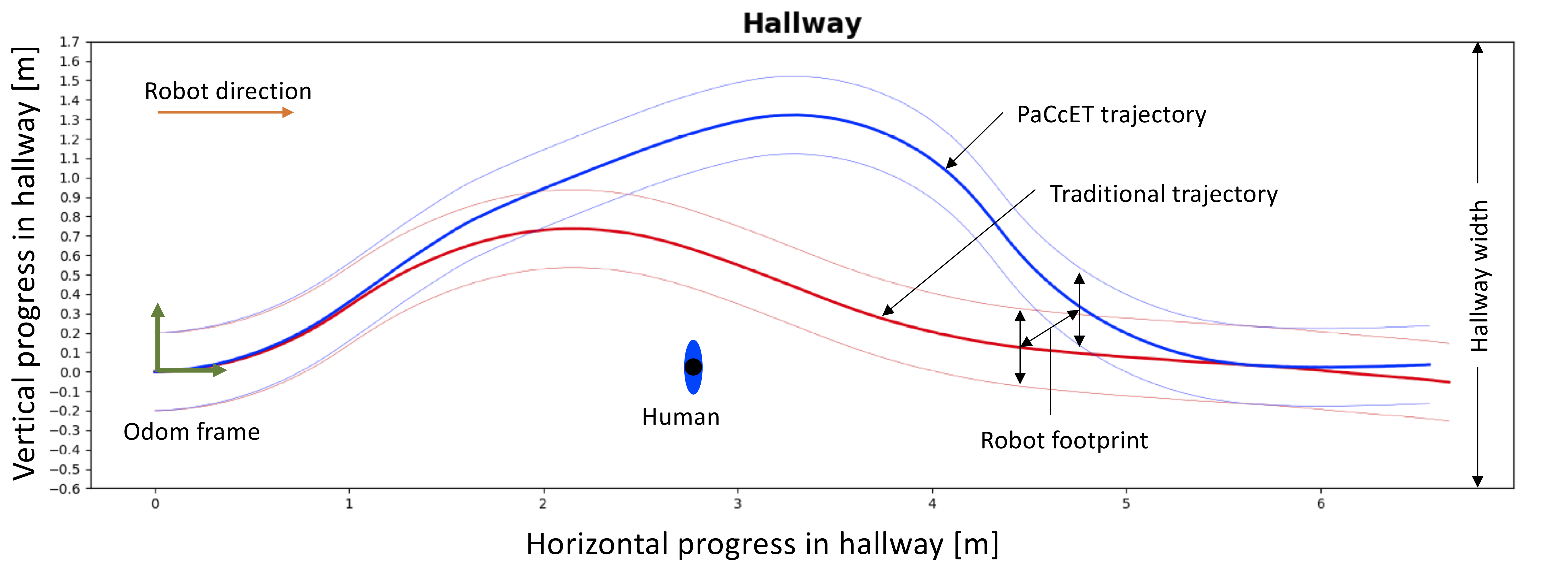}
\caption{Scenario 1: (real-world interaction) Pioneer robot encounters a stationary human standing in the path of the robot in a hallway.}
\label{fig:static_real}
\end{figure}

We extended hallway scenarios to the real-world by implementing our proposed approach on a Pioneer 3DX robot and validating it in both static and dynamic environments. Figure~\ref{fig:static_real} shows a real-world hallway situation where a human is standing in the path of a robot that is attempting to go down the hallway. The robot, when using the traditional planner, treated the human as a mere obstacle and avoided a collision but violated the personal space rule of the human. On the other hand, our approach using PaCcET-based local planning considered the stationary human's personal space using interpersonal distance objective and deviated from the global trajectory in such a way that the personal space rule is obeyed. In Figure~\ref{fig:static_real}, the blue trajectory is generated by our proposed approach, and the traditional approach generates the red trajectory. 

\begin{figure}
\centering
\includegraphics[width=0.9\textwidth]{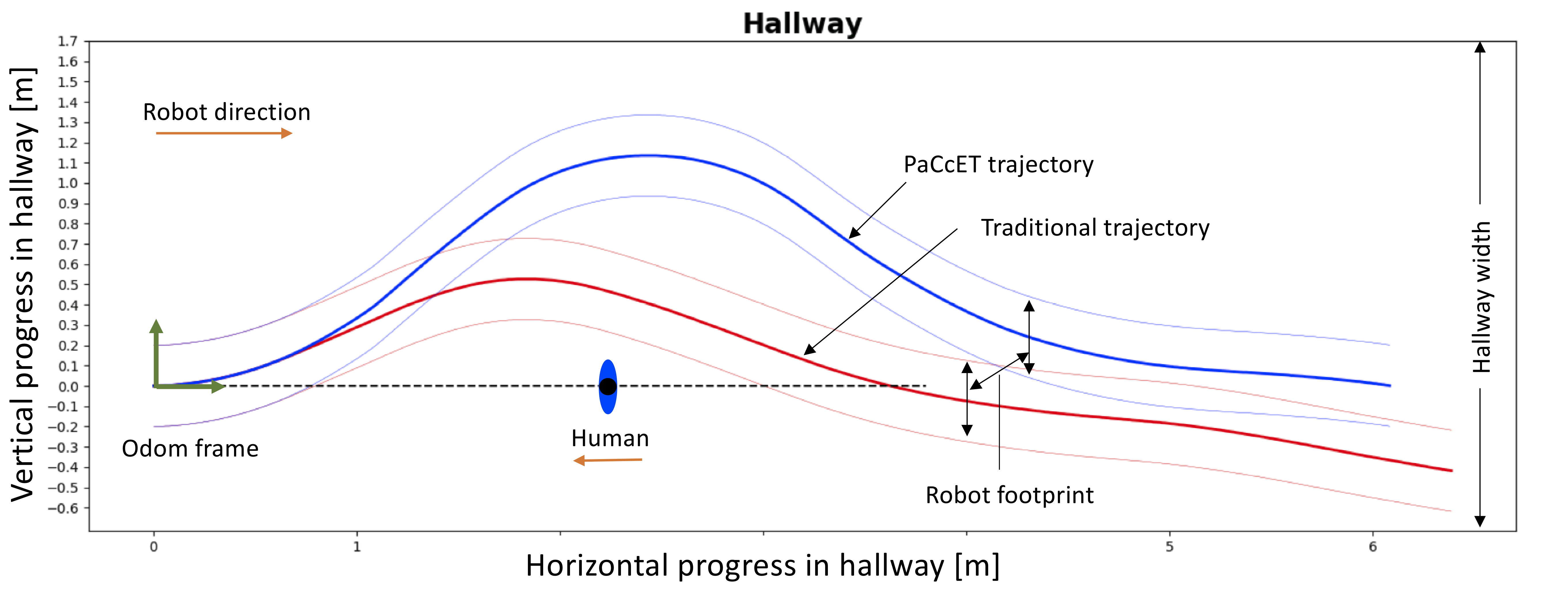}
\caption{Scenario 1: (real-world interaction) Pioneer robot encounter a human walking in the opposite direction and on the side of the hallway.}
\label{fig:dynamic_real}
\end{figure}

Figure~\ref{fig:dynamic_real} shows a real-world hallway interaction like the previous one, but in this case, the human is moving as opposed to a static human. In this experiment, the human is walking in the opposite direction of the robot and also on the wrong side of the hallway. As one can observe, the traditional planner (red trajectory) managed to avoid a collision with the human but went very close to the person, thereby intruding into the human's personal space. Our proposed approach not only avoided a collision but also maintain a safe distance while trying to avoid the human walking on the wrong side of the hallway. Unlike the PR2, Pioneer is a non-holonomic robot; hence, the holonomic behavior, as seen in Figure~\ref{fig:subim4}, is not seen in the real-world interaction.  
It is worth noting that in Figures~\ref{fig:static_real} and \ref{fig:dynamic_real}, the robot with PaCcET trajectory planner showed signs of legibility of movements. In both these cases, the efforts of the robot trying to clear the human's personal space are clearly seen using our method as opposed to the traditional planner.

\subsection{Complex Scenarios}
\begin{figure*}[th]
\centering
\begin{subfigure}{0.47\textwidth}
\centering
\includegraphics[height=4.5cm]{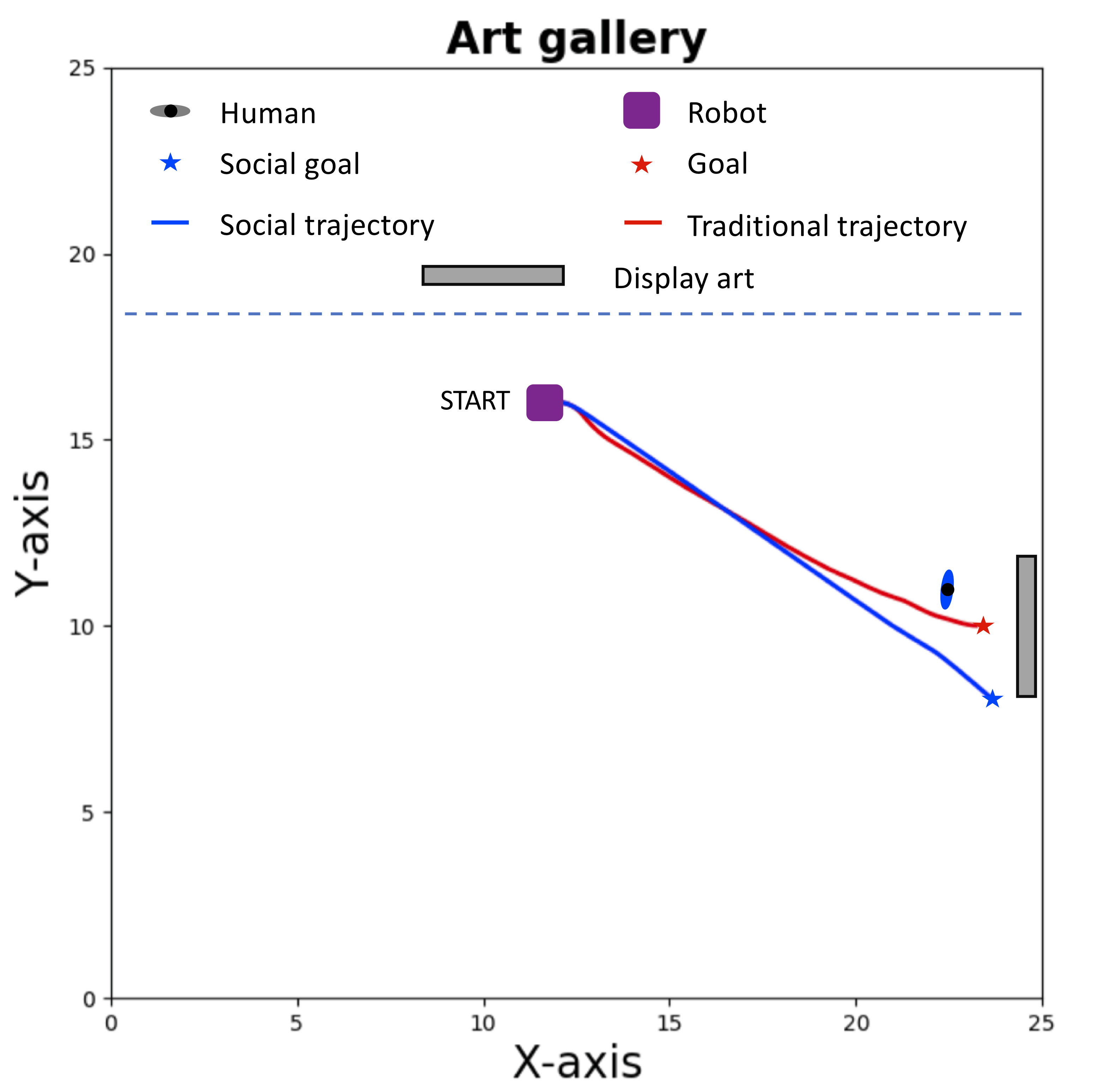} 
\caption{Scenario 2 (simulation): Robot interacting with a human in an art gallery where the robot with SAN planner presents itself at a position appropriate to talk about the art on display, the blue trajectory is generated using the proposed SAN planner.}
\label{fig:art_gallery_1}
\end{subfigure}
\hspace{4mm}
\begin{subfigure}{0.47\textwidth}
\centering
\includegraphics[height=4.5cm]{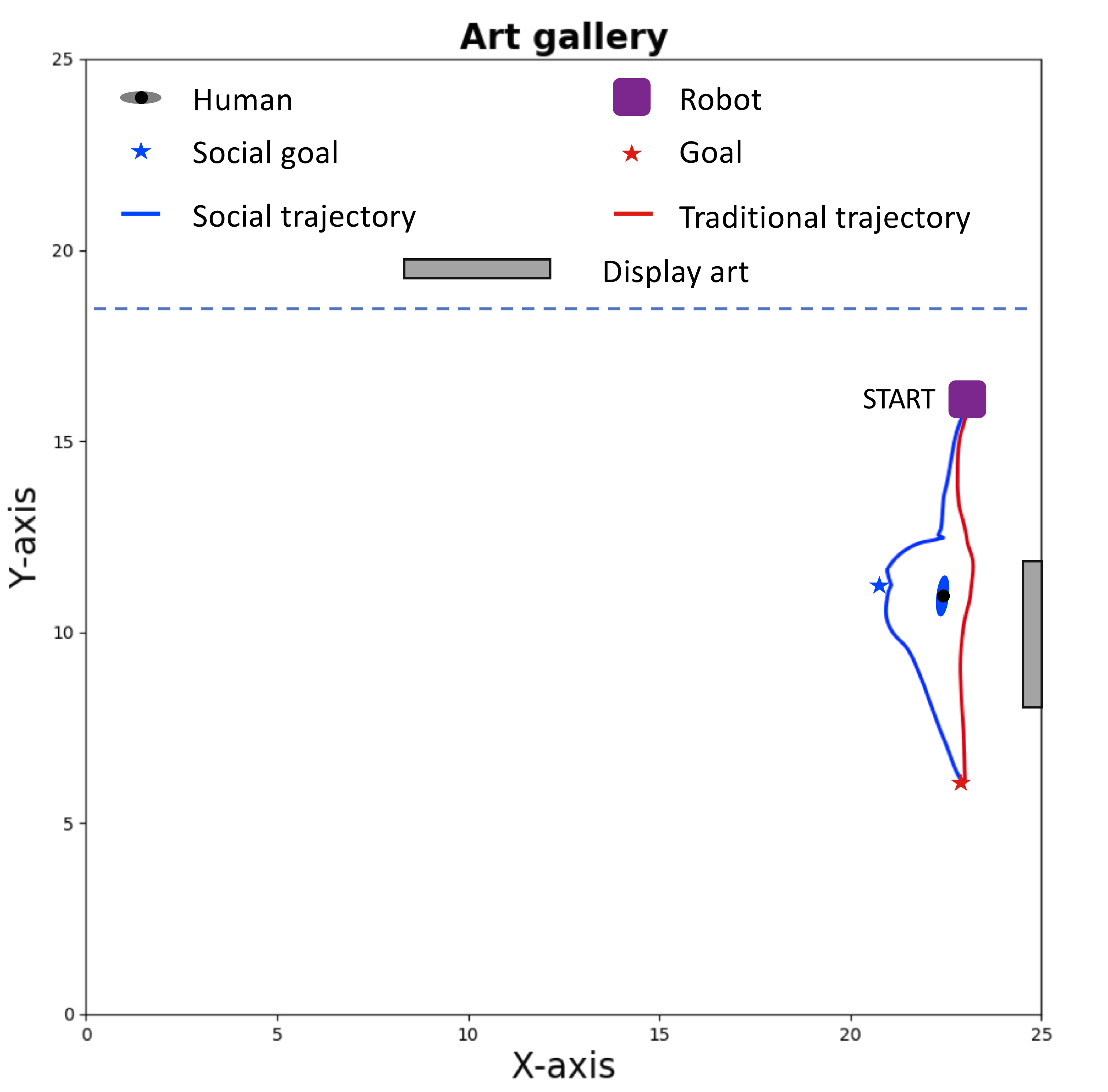}
\caption{Scenario 2 (simulation): Robot taking onto account activity space in an art gallery where the robot with SAN planner avoids going into the activity space, represented by the blue trajectory.}
\label{fig:art_gallery_2}
\end{subfigure}
\hspace{4mm}
\begin{subfigure}{0.47\textwidth}
\centering
\includegraphics[height=4.5cm]{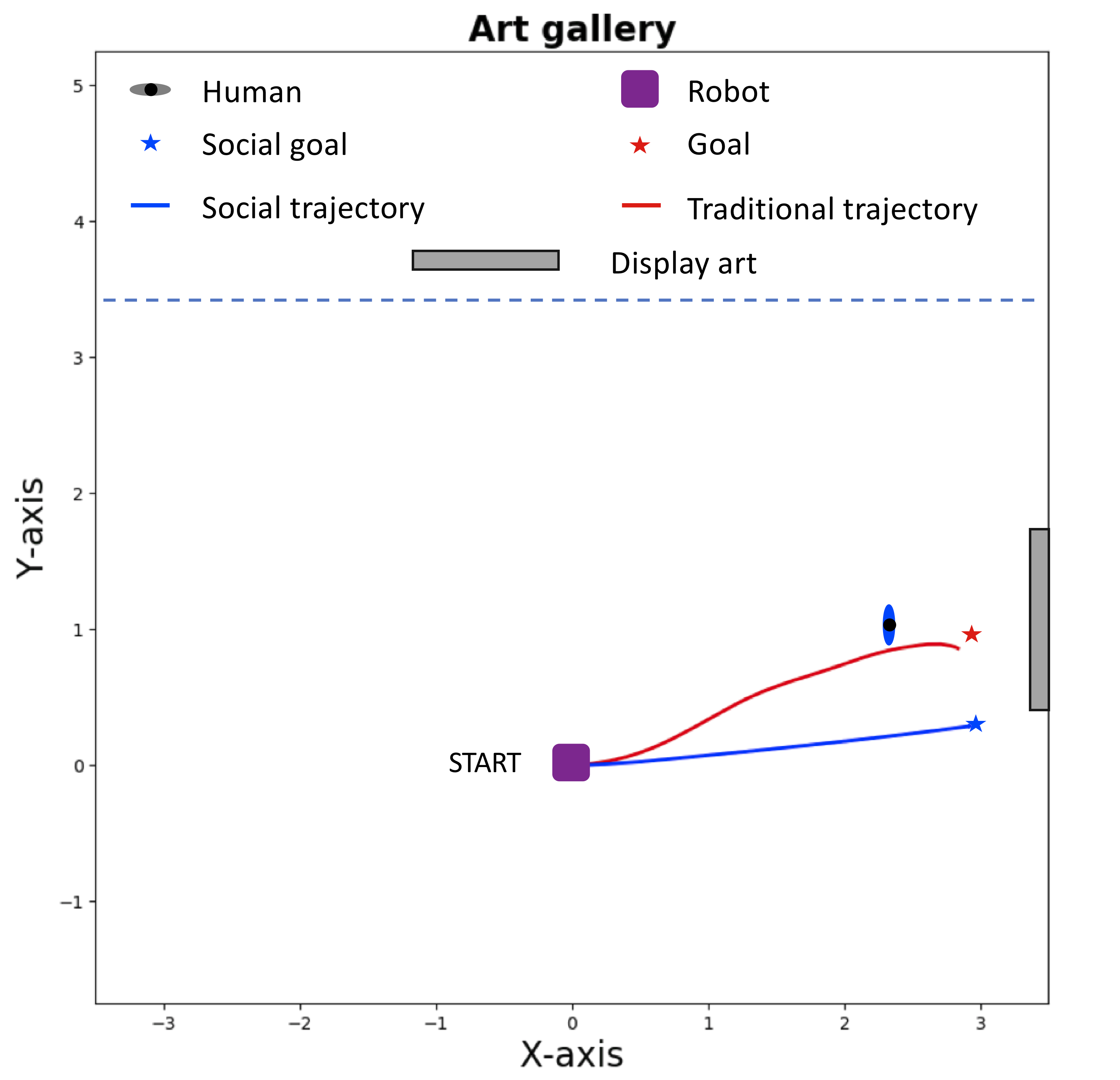} 
\caption{Scenario 2 (real-world): Pioneer robot interacting with a human in an art gallery where the robot with SAN planner presents itself at a position appropriate to talk about the art on display, the blue trajectory is generated using the proposed SAN planner.}
\label{fig:art_gallery_3}
\end{subfigure}
\hspace{4mm}
\begin{subfigure}{0.47\textwidth}
\centering
\includegraphics[height=4.5cm]{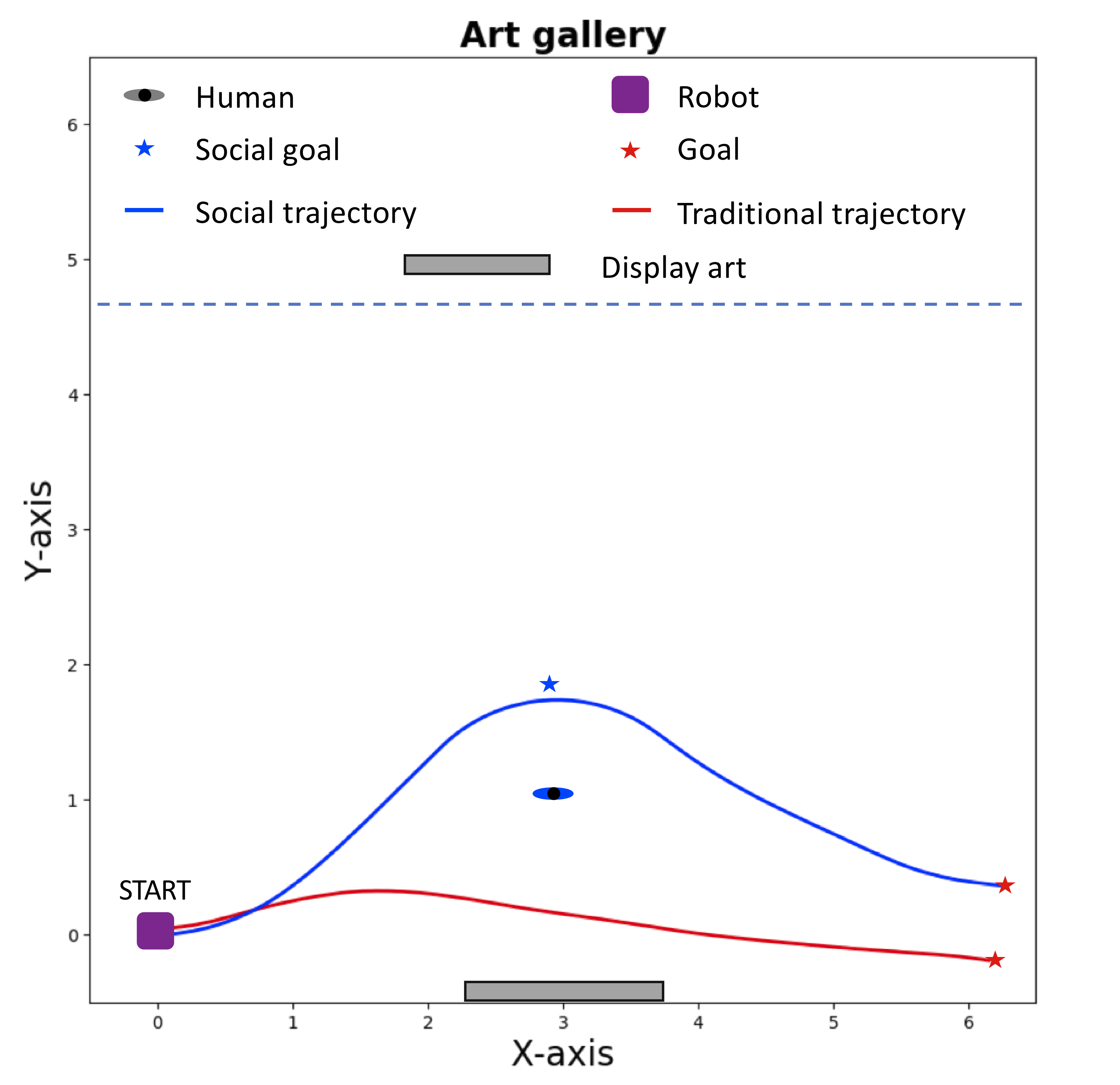}
\caption{Scenario 2 (real-world): Pioneer robot taking onto account activity space in an art gallery where the robot with SAN planner avoids going into the activity space, represented by the blue trajectory.}
\label{fig:art_gallery_4}
\end{subfigure}
\caption{Validation results of Scenario 2 (art gallery) in both simulation and real-world.}
\label{fig:art_gallery}
\end{figure*}

\begin{figure*}[th!]
\centering 
\begin{subfigure}{0.47\textwidth}
\centering
\includegraphics[height = 4.5cm]{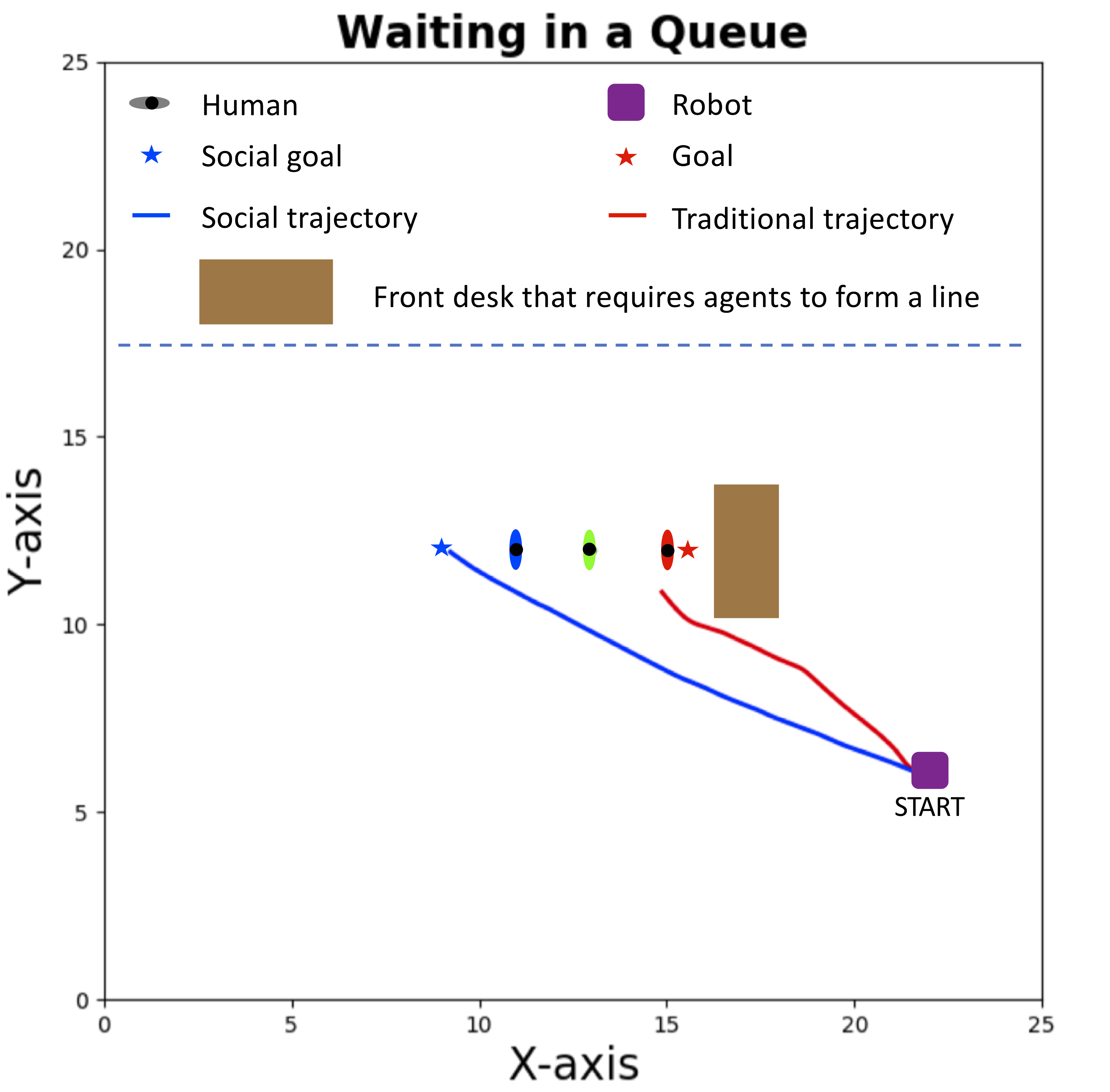}
\caption{Scenario 3: The robot is joining a line, formed in front of a desk. The traditional planner generated the red trajectory, positioned the robot in an inappropriate location beside the first person while attempting to reach the front of the desk. The blue trajectory is generated using our proposed SAN planner leading the robot to join the line (appropriate).}
\label{fig:linesubim1}
\end{subfigure}
\hspace{4mm}
\begin{subfigure}{0.47\textwidth}
\centering
\includegraphics[height=4.5cm]{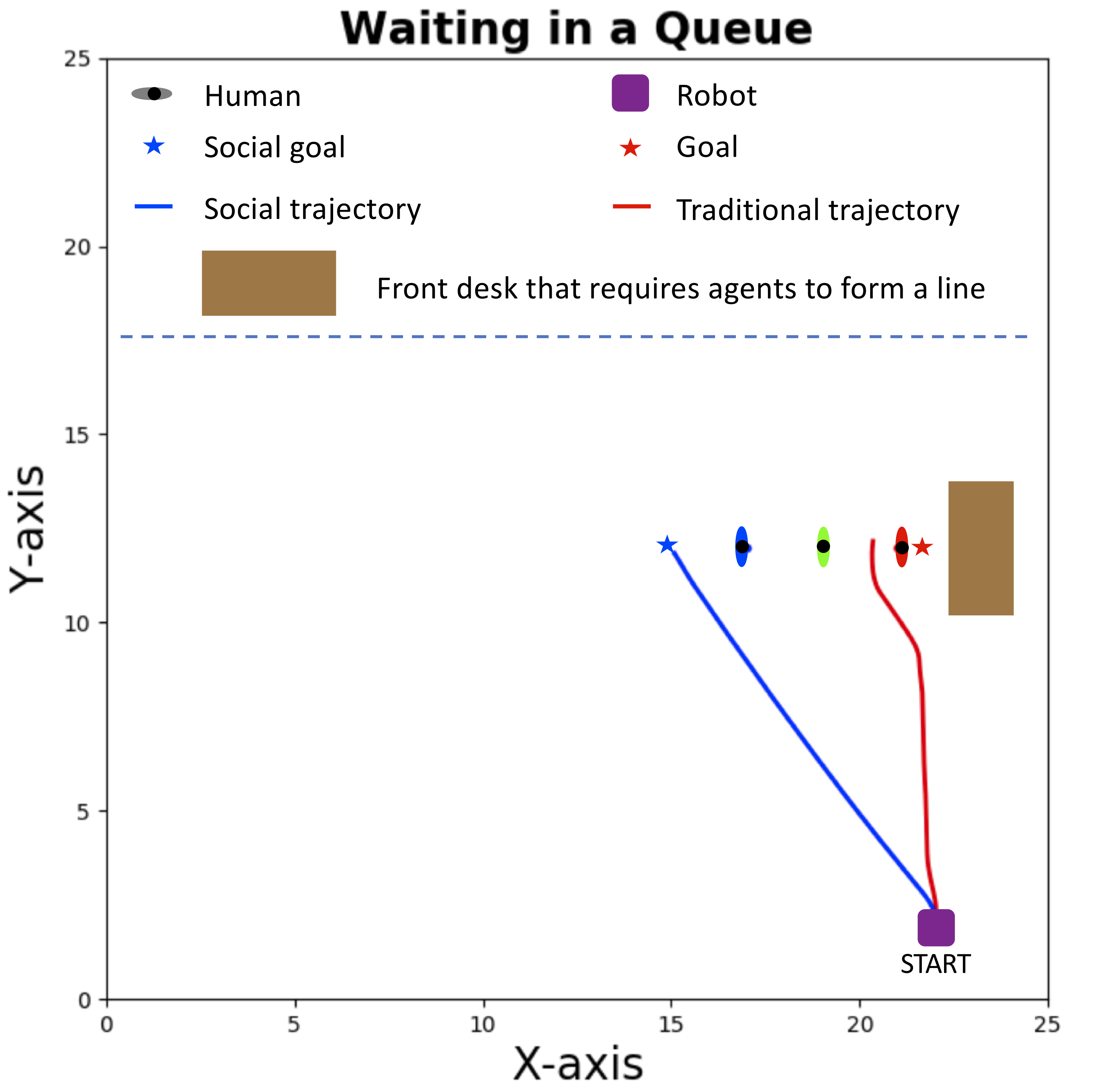} 
\caption{Scenario 3 (location change): The robot is joining a line, formed in front of a desk scenario. The traditional planner generated the red trajectory, guiding the robot between the first two people (inappropriate). The blue trajectory, our proposed approach, leading the robot to join the line (appropriate).}
\label{fig:linesubim2}
\end{subfigure}
\begin{subfigure}{0.47\textwidth}
\centering
\includegraphics[height=4.5cm]{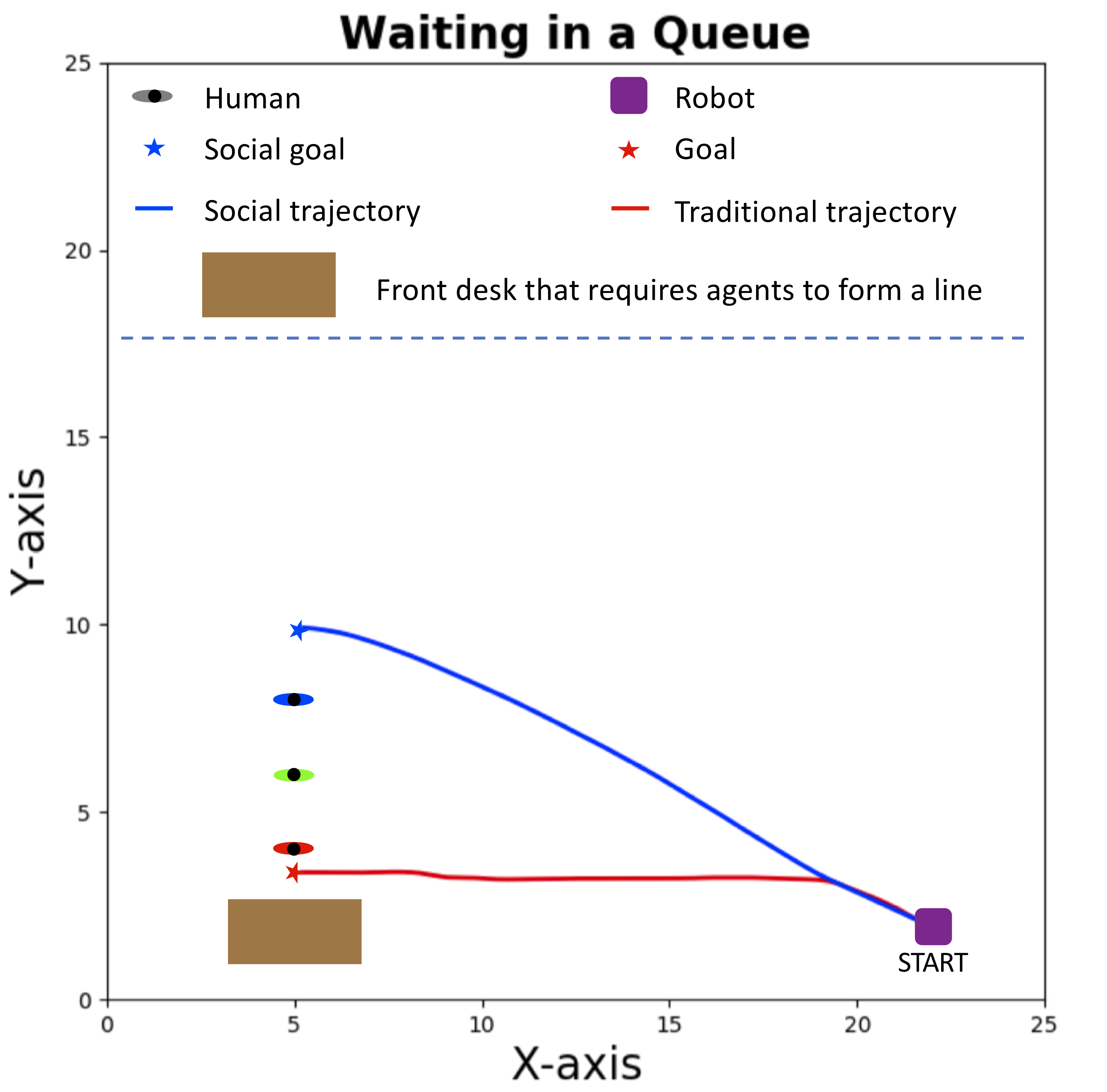}
\caption{Scenario 3 (location and orientation change): The robot is joining a line, formed in front of a desk scenario. The traditional planner generated the red trajectory, guiding the robot to the front of the desk, cutting the line (inappropriate). The blue trajectory, our proposed approach, leading the robot to join the line (appropriate).}
\label{fig:linesubim3}
\end{subfigure}
\hspace{4mm}
\begin{subfigure}{0.47\textwidth}
\centering
\includegraphics[height=4.5cm]{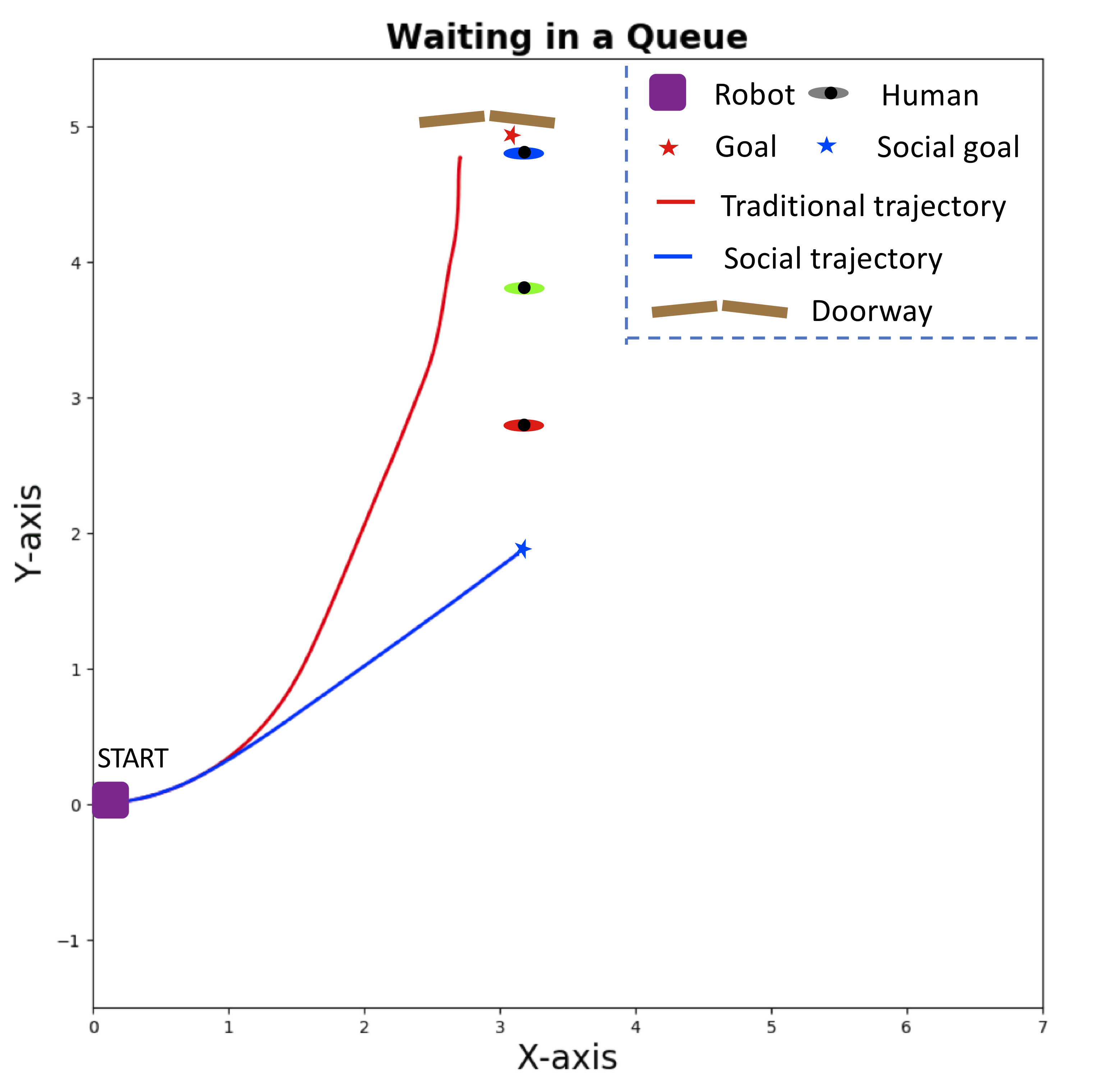}
\caption{Scenario 3 (real-world): Pioneer robot is joining a line, formed in front of a doorway scenario. The traditional planner generated the red trajectory, guiding the robot to a location besides the first person (inappropriate), cutting the line. The blue trajectory, our proposed approach, leading the robot to join the line (appropriate).}
\label{fig:linesubim4}
\end{subfigure}
\caption{Validation results of Scenario 3 (waiting in a queue) in both simulation and real-world.}
\label{fig:line}
\end{figure*}

\begin{figure*}[th!]
\centering 
\begin{subfigure}{0.47\textwidth}
\centering
\includegraphics[height = 4.5cm]{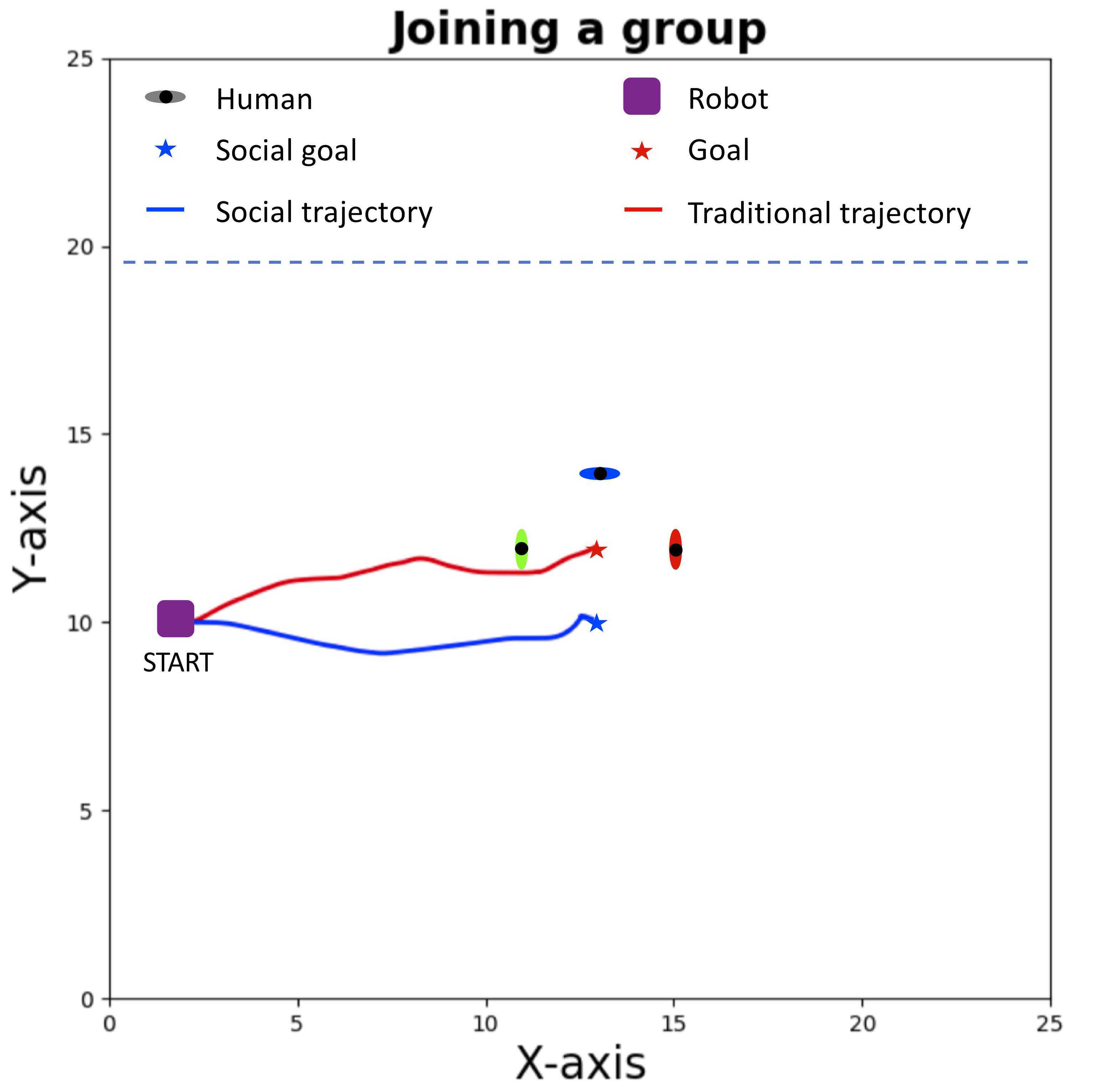}
\caption{Scenario 4: The robot is joining a group where the robot with SAN planner forms an O-formation in order to interact with the group. The traditional planner generates the red trajectory and places the robot in the center of the group. Proposed SAN planner generated the blue trajectory which leads the robot to form an O-formation.}
\label{fig:circlesubim1}
\end{subfigure}
\hspace{4mm}
\begin{subfigure}{0.47\textwidth}
\centering
\includegraphics[height=4.5cm]{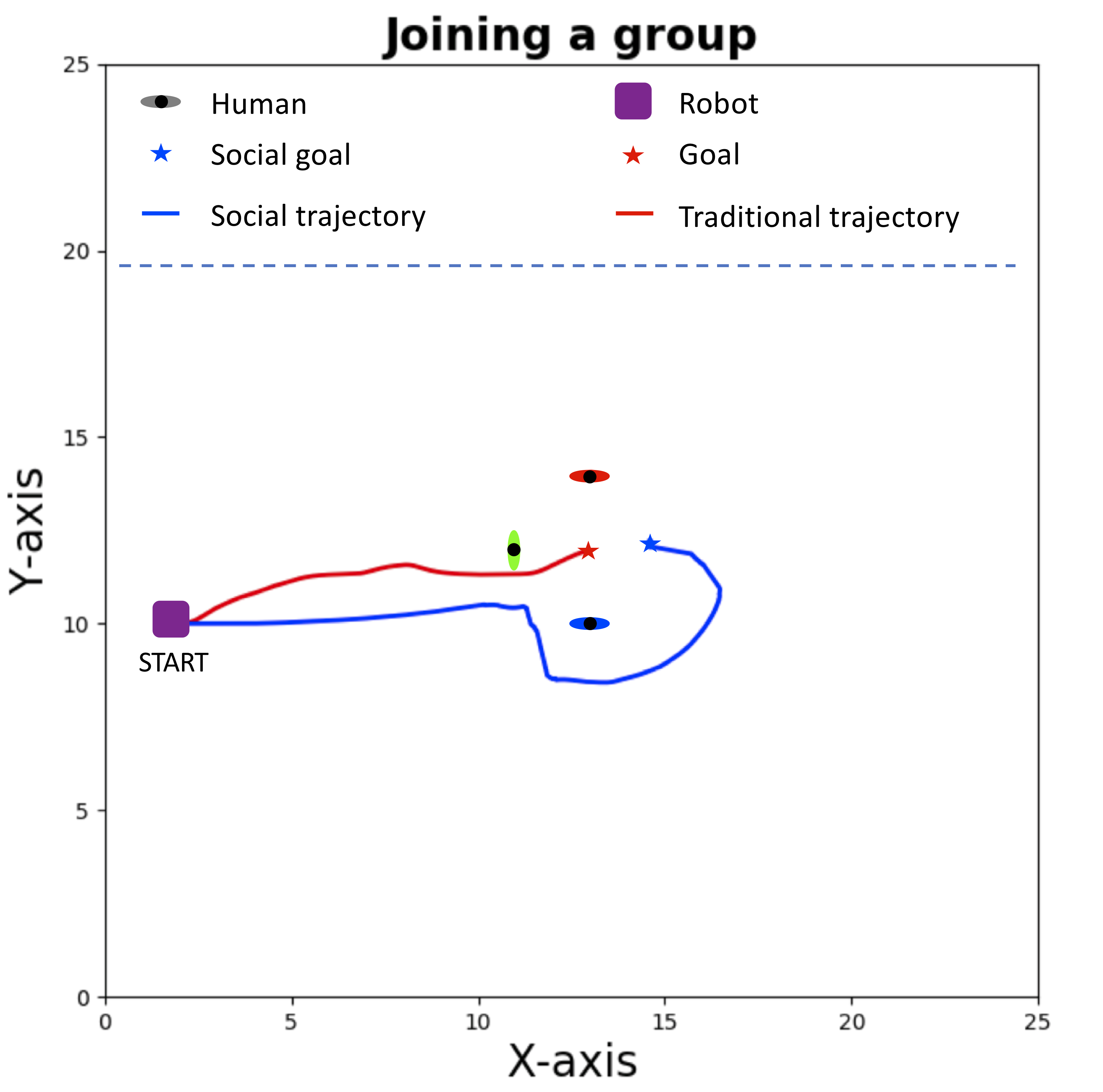}
\caption{Scenario 4 (change in group's open spot):  The traditional planner generated the red trajectory and placed the robot in the center of the group while navigating between two people (inappropriate). Proposed approach generated the blue trajectory which leads the robot to form an O-formation (appropriate).}
\label{fig:circlesubim2}
\end{subfigure}
\begin{subfigure}{0.47\textwidth}
\centering
\includegraphics[height=4.5cm]{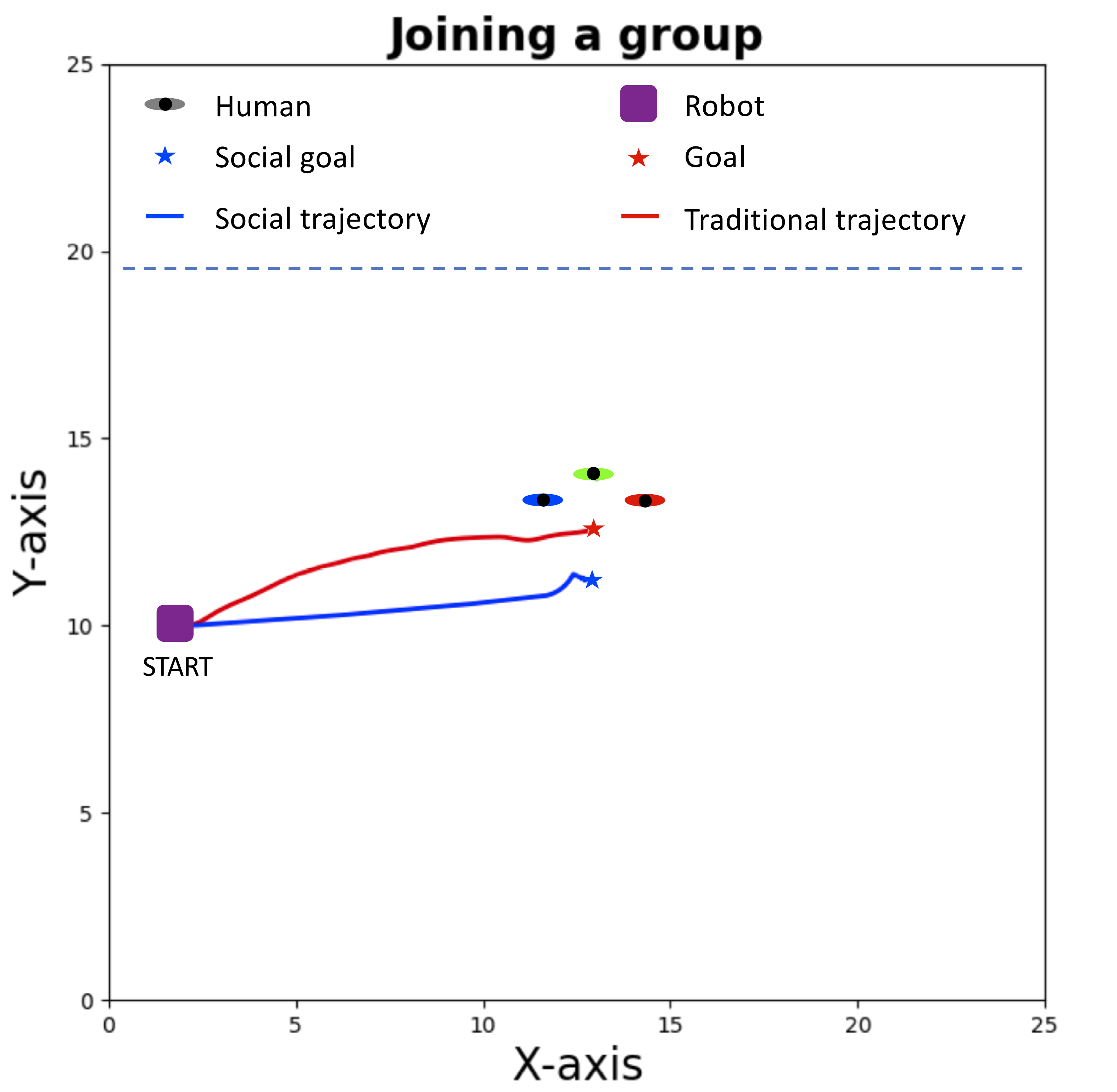}
\caption{Scenario 4 (robot leading the group's conversation): The traditional planner generated the red trajectory and placed the robot in the center of the group. Proposed approach generated the blue trajectory which leads the robot to form an O-formation (appropriate).}
\label{fig:circlesubim3}
\end{subfigure}
\hspace{4mm}
\begin{subfigure}{0.47\textwidth}
\centering
\includegraphics[height = 4.5cm]{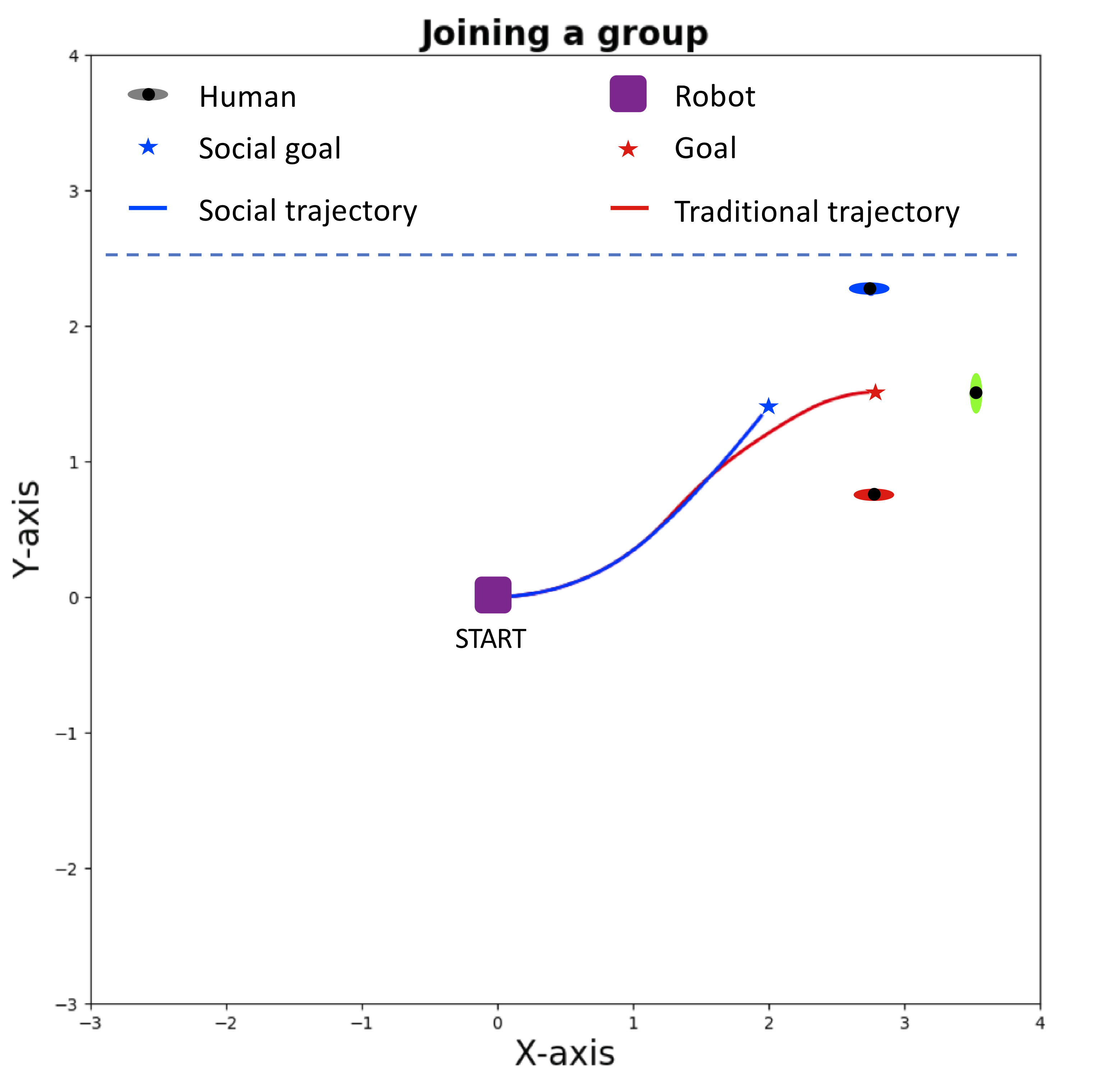}
\caption{Scenario 4 (real-world): Pioneer robot with SAN planner is joining a group, forms an O-formation in order to interact with them. The traditional planner generates the red trajectory and places the robot in the center of the group. Proposed SAN planner generated the blue trajectory which leads the robot to form an O-formation.}
\label{fig:circlesubim4}
\end{subfigure}
\caption{Validation results of Scenario 4 (joining a group) in both simulation and real-world.}
\label{fig:circle}
\end{figure*}

In the previous section, \ref{sec:simple_results}, we showed both in simulation and real-world that by just considering one social feature, i.e., interpersonal distance, our approach was able to account for personal space while navigating a hallway (with different maneuvers of a human partner). In this section, we will see the results of our approach applied to complex social scenarios like \textit{art gallery interactions} (Figure~\ref{fig:art_gallery}), \textit{waiting in a line} (Figure~\ref{fig:line}) and \textit{joining a group of people} (Figure~\ref{fig:circle}). These scenarios are representative of both human-human and human-environment interactions that occur in normal social discourse. 

Figure \ref{fig:art_gallery_1} shows the behavior of our social planner and traditional planner in an \textit{art gallery} situation (three objectives) in simulation. We considered an art gallery scenario, but it can be generalized to other similar scenarios like a tour guide robot in a museum or an attraction. For this scenario, we staged a human-robot interaction consisting of a robot presenting a piece of art (hanging to a wall) to a human standing nearby. Both the traditional planner (red line) and the SAN planner (blue line) were given the same goal (represented as a red star) and start (START) locations. The traditional planner steered the robot to the goal location, cutting the standing person from the back (inappropriate). On the other hand, the SAN planner steered the robot to a location that is appropriate to present the details of the art to the human (social goal). The SAN planner approached the social goal, leaving enough personal space based on the interpersonal distance feature. 

Art gallery interactions are not always presenting the artwork on display. While navigating an art gallery, one should consider the affordance and activity spaces between the artwork and an individual looking at the art. Activity space is a social space linked to actions performed by agents~\cite{lindner2011towards}. For example, the space between the subject and a photographer is an activity space, and we humans generally avoid getting in the way of such activity spaces. Affordance space is defined as a social space related to a potential activity provided by the environment~\cite{rios2015proxemics}. In other words, affordance spaces are potential activity spaces. An environment like an art gallery provides numerous locations as affordance spaces (place in front of every piece of art is an affordance space). When a visitor steps into one such affordance space, that space between the artwork and the interacting human becomes activity space.

In Figure~\ref{fig:art_gallery_2}, we demonstrated an appropriate behavior around activity space in simulation using our proposed SAN planner. For this scenario, we staged a human-robot interaction consisting of a human interacting with a piece of art working hanging to the wall. Both the traditional planner (red line) and the SAN planner (blue line) were given the same goal (represented as a red star) and start (START) locations. The traditional planner steered the robot to the goal but did not account for the activity space, i.e., the robot traversed through the activity space (inappropriate). On the other hand, the PaCcET-based SAN planner steered the robot to the goal location while avoiding the activity space (appropriate social behavior). The social goal, while avoiding an activity zone, is not an end goal where the robot would stop but is more like a social goal that acts as a way-point in reaching the end goal.

Similarly, the two art gallery behaviors (presenting art and avoiding activity space) is implemented and validated on a Pioneer robot, and the results are shown in Figure~\ref{fig:art_gallery_3} and Figure~\ref{fig:art_gallery_4}

Figure \ref{fig:linesubim1} shows the behavior of our social planner and traditional planner in the \textit{waiting in a queue} situation (five objectives) in simulation. Here, we considered a front desk interaction, but this can be generalized to other similar social scenarios where a robot or a human is required to form a line before reaching the goal. For example, social scenarios like getting coffee from a public coffee machine, taking an elevator, etc. In this context, we staged a human-robot interaction consisting of a robot that wants to interact with a front desk representative of an office building where other people were being served on a first-come-first-served basis. Both the traditional planner (red line) and the SAN planner (blue line) were given the same goal (represented as a red star) and start (START) locations. The trajectory planner tried to steer the robot to the goal location and stopped at an inappropriate location (besides the person currently being served) as the traditional planner treated the human as an object. On the other hand, the SAN planner steered the robot to an appropriate location, i.e., end of the line positioning the robot behind the last person (social goal), considering personal space as well. 

Figure \ref{fig:linesubim2} and \ref{fig:linesubim3} shows results with variations in scenarios 3 (\textit{waiting in a queue}). The variations are the locations of people and the orientation of the queue formed by them, figures \ref{fig:linesubim2} and \ref{fig:linesubim3} show that our method is robust. Figure \ref{fig:linesubim1} shows the behavior of our social planner and traditional planner in the \textit{waiting in a queue} situation (five objectives) in the real-world. Here, considered a doorway social situation where we humans expect to go one after the other and not rush or cut the line. Both the traditional planner (red line) and the SAN planner (blue line) were given the same goal (represented as a red star) and start (START) locations. The traditional trajectory planner tried to steer the robot to the goal location and stopped at an inappropriate location (besides the first person in front of the door). On the other hand, the SAN planner steered the robot to an appropriate location, i.e., end of the line positioning the robot behind the last person (social goal), considering personal space as well. 

Figure \ref{fig:circlesubim1} shows the behavior of our social planner and traditional planner in \textit{Joining a group} situation in simulation. Here, we considered an HRI situation where the robot is required to join a group of three people. However, this can be generalized to interact with more people. Both the traditional planner (red line) and the SAN planner (blue line) were given the same goal (represented as a red star) and start (START) locations. The trajectory planner steered the robot to position it at an awkward location (middle of an interacting group) as the traditional planner did not account for group proxemics and group dynamics. On the other hand, the SAN planner steered the robot to an appropriate location, i.e., a vacant spot on the circle, considering group proxemics.

Figure \ref{fig:circlesubim2} shows a variation in the open spot where the robot needs to join. In this case, the open spot is in a tricky location as the robot has to approach the group from the back. Our proposed approach found a way around the group to the social goal. On the other hand, the traditional planner leads the robot to the group's center while getting in between two people (blue and green shirts).

Figure \ref{fig:circlesubim3} not only differs in the size of the circle formed but also is a variation of O-formation where the robot is leading the conversion opposing to joining a conversation. When joining a group for discussion, we tend to maintain a uniform spacing between every member of a group wherein joining a group to lead the conversation, all the members of the group except the lead squish together so that the leader can make eye contact with all the members (with as little field of view as possible). In this case, the traditional planner guides the robot to the center of the group, whereas the proposed method guides the robot to a social goal location where the robot can effectively interact with the group.

Figure \ref{fig:circlesubim4} shows the results of our method implemented on a Pioneer robot executing an appropriate social behavior of \textit{joining a group}.  Both the traditional planner (red line) and the SAN planner (blue line) were given the same goal (represented as a red star) and start (START) locations. The trajectory planner steered the robot to position it at an inappropriate location (the middle of an interacting group) as the traditional planner did not account for group proxemics and group dynamics. On the other hand, the SAN planner steered the robot to an appropriate location, i.e., a vacant spot on the circle, considering group proxemics.

\subsection{Comparison Results}
We compare our optimization-based SAN approach with a recent inverse reinforcement learning-based SAN method~\cite{kollmitz20iros}. Kollmitz \textit{et al.} \cite{kollmitz20iros} investigated force control as a natural means to teach social trajectories to robots using inverse reinforcement learning trained using demonstrations. The IRL method learned hallway passing through demonstrated trajectories. We used the hallway passing scenario as a common task for all three planners, and the results for various metrics are shown in table~\ref{table:metrics} (Average of 5 runs). The simulated task involves a robot passing a human in a hallway, the length of the hallway scenario (straight line distance) is kept at 5 meters, and the robot meets a simulated human halfway through the hallway. 
We used the following metrics to compare our work:

\begin{enumerate}
    \item \textbf{Total distance:} Total distance traveled by the robot to reach the goal.
    \item \textbf{Time to reach goal:} Total time taken to reach the goal location.
    \item \textbf{Lateral distance maintained:} The lateral distance maintained by the robot when passing the human.
    \item \textbf{No. of Proxemic intrusions:} Number of times the robot intruded into the human's personal space.
    \item \textbf{Distance when robot deviated:} Distance from the human when the robots starts to deviate. This metric may contribute to the legibility and predictability of a robot trajectory~\cite{banisetty2018towards, dragan2013legibility}. 
\end{enumerate}

\begin{table}[h!]
\centering
\begin{center}
 \begin{tabular}{||c c | c c c||} 
 \hline
 Metric & Expected & Traditional & IRL & PaCcET \\ [0.5ex] 
 \hline\hline
 Total distance \textbf{($D_t$)} & LOW & \textbf{5.04m} & 5.21m & 5.34m \\ [0.5ex]
 \hline
 Time to reach goal \textbf{($t$)} & LOW & \textbf{13.2s} & 15.94s & \textbf{14.36s} \\ [0.5ex]
 \hline
 Lateral distance maintained \textbf{($D_l$)} & HIGH & 0.9m & 1.01m* & \textbf{1.39m} \\ [0.5ex]
 \hline
 No. of Proxemic Intrusions \textbf{($N$)} & LOW & 5 & \textbf{0}* & \textbf{0} \\ [0.5ex]
 \hline
 Distance when robot deviated \textbf{($D_d$)} & HIGH & n/a & 1.8m & \textbf{1.5m} \\ [0.5ex]
 \hline
\end{tabular}
\end{center}
\caption{Comparison of \textit{traditional navigation}, \textit{IRL based social navigation}, and \textit{our approach} on the \textit{hallway passing} scenario (results are average of 5 runs). *The number of proxemic intrusions into personal space is considered 0 for IRL based method even though the IRL approach maintained an average lateral distance of 1.01 m only (the max range of personal space according to Hall's model of proxemics is 1.2 meters). This is because the participants trained the robot for their comfortable distance; hence, we assume the robot did not intrude the personal zone (Proxemic theory depends on many factors such as familiarity with robots).}
\label{table:metrics}
\end{table}

It is evident from Table~\ref{table:metrics} that the traditional planner outperformed other social navigation methods in terms of performance metrics such as \textit{distance traveled} and \textit{time taken to reach the goal}. On the other hand, traditional navigation methods under-performed in terms of social metrics such as \textit{lateral distance maintained}, \textit{proxemic intrusions}, and \textit{distance from a human when the robot started to show deviation in its path}. However, it is worth noting that it is acceptable for a social robot to choose a performance-based sub-optimal path as the quality of interaction takes precedence over performance. The distance traveled by the IRL-based SAN method is less than the distance traveled by our approach, and the difference is explained by the fact that our approach maintained a larger \textit{lateral distance} when passing a person. On the \textit{time to reach goal} metric, our method performed better as our method is a modification of traditional navigation.

On social metrics, our method performed well on \textit{lateral distance maintained} compared to the IRL and traditional navigation methods. Our experiments show that our approach and IRL-based approach reported zero proxemic intrusions into personal space (max range of 1.2 meters). However, it is worth noting that a person's comfortable personal space with a robot depends on numerous factors (both robot and human) such as gender, familiarity with robots~\cite{takayama2009influences}. The number of proxemic intrusions into personal space is considered zero for IRL based method even though the IRL approach maintained an average lateral distance of 1.01m only (the max range of personal space according to Hall's model of proxemics~\cite{hall1966hidden} is 1.2 meters). This is because the participants trained the robot for their comfortable distance; hence, we assume the robot did not intrude the personal zone. On the other social metric (\textit{distance from which the robot started to deviate}), the IRL based planner performed slightly better when compared to our approach, and we attribute this to the fact that the IRL based social navigation is a global planner, whereas our approach to social navigation is at the local planning stage. Planning for social objectives with global planning happens at the beginning of the planning task, whereas planning for social objectives at a local planning stage happens as the robot is approaching a person. Hence, the IRL-based planner was able to deviate when passing the person much earlier than our approach. It is worth noting that both the social navigation methods outperformed the traditional planner as the traditional planner did not even deviate from its planned trajectory as it passed the human very closely. Such deviations before passing a person may lead to legibility and predictability of trajectories~\cite{dragan2013legibility}.  


\section{Discussion and Future Work}
\label{sec:discussion}

With a series of experiments in simulation and with a real robot using a multi-objective optimization tool like PaCcET at the local planning stage of navigation, we showed that social norms related to proxemics could be addressed in a SAR system which in turn can aid the acceptance of robots in human environments. We showed that our approach could handle a simple, single-person interaction in a hallway scenario. We also showed that our approach could handle sub-scenarios (different types of interaction in a scenario) such as a stationary human in a hallway, passing behavior, etc. Similarly, in an art gallery situation, our methods showed that it is applicable for sub-scenarios like presenting artwork and avoiding activity zones. We demonstrated the generalizability of our approach by introducing multiple humans in complex social scenarios with multiple features like interpersonal distances, group proxemics, activity zone, and social goal distance (adherence to social goal). Finally, we showed that our modified local planner could adjust to changes in the scenarios like the location of people, different line formations, and O-formations. 


This work dealt with low-level decisions of which future trajectory points are better for the given interaction scenario. One assumption in this work is that the robot has prior knowledge of the on-going interaction. We will extend this work using a knowledge graph approach~\cite{shahrezaie2021towards}, implementing a high-level decision-making system that can select the crucial objectives for the quality of HRI for an autonomously sensed scenario~\cite{banisetty2018towards}. We will extend this work by validating this system using its conformity to social metrics defined by the social parameters we discussed and surveying the perception of social intelligence of the resultant behavior. We plan to utilize not just the distance-based feature but also features related to orientation like heading angle, heading difference between the robot and the people/group, etc., along with features related to the environment, such as the position of the agents (robot and people) in the environment (for example, in a hallway, distance from the right side of the hallway)~\cite{sebastian2017socially-aware}.

When examining the social impact of a SAN system, it is important that any instruments used are properly assessing social intelligence. Kruse \textit{et al.} \cite{kruse2013human} identified Comfort, Sociability, and Naturalness as challenges that SAN planners should tackle in a collaborative human environment \cite{kruse2013human}. We identified other challenges like predictability, legibility, safety, acceptance, etc., and working on providing clear definitions and metrics/methods for measuring them. Perceived Social Intelligence (PSI) is an important parameter we identified, which has importance in robot motion in human environments. Social intelligence is the ability to interact effectively with others to accomplish one's goals \cite{ford1983further}. Social intelligence is critically important for any robot that will be around people, whether engaged in social or non-social tasks. Some aspects of robotic social intelligence have been included in HRI research \cite{moshkina2012reusable, bartneck2009measurement, nomura2006measurement, ho2010revisiting}, but current measures are brief and often include extraneous variables. We designed a comprehensive instrument for measuring the PSI of robots~\cite{kim2018psi, barchard2019perceived, barchard2020measuring}, which should more precisely measure the social impact of our approach on people in the robot's environment and people observing those interactions~\cite{banisetty2021implicit, honour2021perceived}.

\section{Conclusion}
\label{sec:conclusion}
We presented a novel approach to the socially-aware navigation (SAN) problem at a local planning stage by transforming the Pareto front to an objective space (forced to be convex) using Pareto Concavity Elimination Transformation (PaCcET) method. PaCcET was implemented in local planning of well established ROS navigation stack to deal with spatial communication at a low-level planning stage. We validated the developed system both in simulation and on a mobile robot to show the applicability of our proposed approach with multiple scenarios involving multiple humans.
A follow-up study will investigate the social aspect of the navigation behaviors using scales that already exist and new scales that our group is currently investigating.

\section*{Acknowledgment}
The authors would like to acknowledge the financial support of this work by the National Science Foundation (NSF, \#IIS-1719027), Nevada NASA EPSCoR (\#NNX15AI02H), and the Office of Naval Research (ONR, \#N00014-16-1-2312, \#N00014-14-1-0776). We would like to acknowledge the help of Vineeth Rajamohan, Fausto Vega, Ashish Kasar, Athira Pillai and Andrew Palmer.

%
\bibliographystyle{ACM-Reference-Format}
\bibliography{bibo}

%
\appendix

\end{document}